%% file: main.tex
\def\arxivbuild{}
\def\revisedblack{}
\documentclass[runningheads]{llncs}

\usepackage[year=2026]{accv}

\usepackage{accvabbrv}      
\usepackage{graphicx}
\usepackage{booktabs}
\usepackage{multirow}
\usepackage{colortbl}
\usepackage{amsmath}
\usepackage{amssymb}
\usepackage{tikz}
\usetikzlibrary{arrows.meta,positioning,fit,backgrounds}
\usepackage[accsupp]{axessibility}  
\usepackage{placeins}

\usepackage[breaklinks,colorlinks,citecolor=accvblue]{hyperref}
\usepackage{orcidlink}
\usepackage{pifont}   

\input{tex/macros}
\ifdefined\revisedblack\long\def\revised#1{#1}\fi

\begin{document}

\title{Physics-Guided Flow-Map Matching\\ for Precipitation Nowcasting}
\titlerunning{Physics-Guided Flow-Map Matching for Nowcasting}

\author{Shunya~Nagashima\inst{1}\orcidlink{0009-0007-9741-7628}\textsuperscript{(\ding{41})} \and
Takumi~Bannai\inst{2,3}\orcidlink{0000-0002-6827-488X} \and
Makoto~Misaizu\inst{1,4}\orcidlink{0009-0001-4254-1652} \and
Keisuke~Maeda\inst{4}\orcidlink{0000-0001-8039-3462} \and
Takahiro~Ogawa\inst{4}\orcidlink{0000-0001-5332-8112} \and
Miki~Haseyama\inst{4}\orcidlink{0000-0003-1496-1761}}
\authorrunning{S. Nagashima et al.}
\institute{Neurogica Inc., Tokyo, Japan\\
\email{shunya.nagashima@neurogica.com} \and
LTS, Inc., Tokyo, Japan \and
ME-Lab Japan, Inc., Tokyo, Japan \and
Hokkaido University, Sapporo, Japan}

\maketitle

\begin{abstract}
  Precipitation nowcasting, generating future radar fields from past
  observations, is critical for flood warning and disaster response. It is
  also a demanding benchmark for spatiotemporal generative modeling, with
  chaotic dynamics, heavy-tailed intensities, and rare high-intensity
  structures that matter most. Deterministic models
  minimize a pixel loss and are driven toward the conditional mean, which blurs
  exactly those structures, while generative models that add a stochastic
  residual on top of a deterministic backbone inherit the same blur.
  We propose Physics-Guided Flow-Map Matching (PG-FMM), a conditional flow-map
  model that decouples predictable advection from uncertain small-scale detail. A
  frozen Lagrangian advection prior transports the radar field and supplies an
  explicit motion forecast, and a flow-map generative head, conditioned on
  the past frames and the prior rollout rather than summed onto it, produces
  sharp stochastic detail in four sampling steps. The prior serves only as
  guidance, so the head replaces blurred structure instead of inheriting it.
  Extensive experiments on four radar benchmarks show that PG-FMM outperforms
  state-of-the-art methods on $18$ of $24$ metrics, with the largest gains at
  heavy-rain thresholds, where the critical success index improves by up to
  $58.9\%$.
  The project page can be found at \url{https://neurogica.github.io/PG-FMM}.
\end{abstract}


\section{Introduction}
\label{sec:intro}
\input{tex/sections/01_introduction}

\section{Related Work}
\label{sec:related}
\input{tex/sections/02_related_work}

\section{Preliminaries}
\label{sec:prelim}
\input{tex/sections/03_preliminaries}

\section{Method}
\label{sec:method}
\input{figures/overview}
\input{tex/sections/04_method}

\section{Experiments}
\label{sec:exp}
\input{tex/sections/05_experiments}

\section{Conclusion}
\label{sec:conclusion}
\input{tex/sections/06_conclusion}


\bibliographystyle{splncs04}
\bibliography{main}

\ifdefined\arxivbuild
\clearpage
\section*{Supplementary Material}
\input{tex/sections/supp_body}
\fi

\end{document}

%% file: tex/macros.tex
\newcommand{\method}{PG-FMM\xspace}                          
\newcommand{\methodfull}{Physics-Guided Flow-Map Matching\xspace}

\newcommand{\warpop}{\operatorname{warp}}
\newcommand{\clipop}{\operatorname{clip}}
\newcommand{\Expect}{\mathbb{E}}
\newcommand{\Reals}{\mathbb{R}}
\newcommand{\flowmap}{\Phi}                                 
\newcommand{\Tin}{T_{\mathrm{in}}}
\newcommand{\Tout}{T_{\mathrm{out}}}

\newcommand{\Rp}{R^{\text{prior}}}

\long\def\revised#1{#1}   

\makeatletter
\def\institutename{\par
 \begingroup
 \parskip=\z@
 \parindent=\z@
 \setcounter{@inst}{1}%
 \def\and{\par\vspace{1.5pt}\stepcounter{@inst}%
 \noindent$^{\the@inst}$\,\ignorespaces}%
 \setbox0=\vbox{\def\thanks##1{}\@institute}%
 \ifnum\c@@inst=1\relax
   \gdef\fnnstart{0}%
 \else
   \xdef\fnnstart{\c@@inst}%
   \setcounter{@inst}{1}%
   \noindent$^{\the@inst}$\,%
 \fi
 \ignorespaces
 \@institute\par
 \endgroup}
\makeatother

%% file: tex/sections/01_introduction.tex
\begin{figure}[t]
  \centering
  \includegraphics[width=\linewidth]{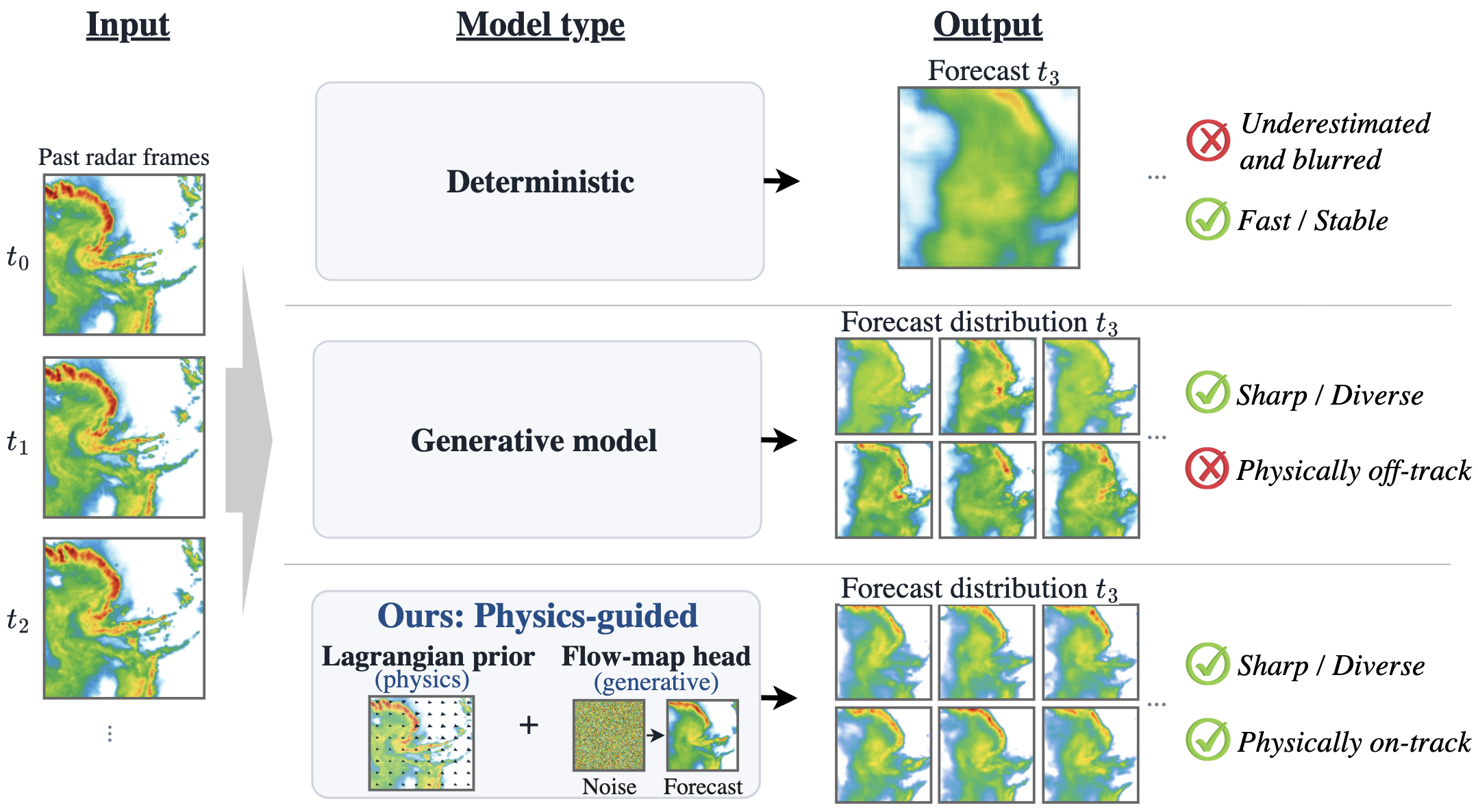}
  \caption{Three families of precipitation nowcasters: (1)~deterministic models
  blur high-intensity cells; (2)~purely generative models can drift off the
  observed storm track; (3)~our physics-guided design conditions a flow-map head
  on a frozen advection prior, keeping forecasts sharp and on track.}
  \label{fig:teaser}
\end{figure}
Precipitation nowcasting generates the next one to two hours of high-resolution
reflectivity fields from a short history of radar frames. It is conditional
video prediction in a demanding regime, as the dynamics are chaotic, the
intensity distribution is heavy-tailed, and forecast value is concentrated in
rare high-intensity structures. Nowcasts underpin
flood early warning, disaster response, water management, and the safe operation
of transport and energy
infrastructure~\cite{survey_nowcasting,dgmr,nowcastnet}, and their importance
grows as a warming climate intensifies short-duration convective rainfall
extremes and the flash flooding they cause~\cite{fowler2021}; the short lead
time of a nowcast often separates an effective response from a costly one. A
useful nowcast must therefore forecast where and how strongly convective cells
will develop, not merely whether it will rain.

This high-intensity regime is also the hardest, because small-scale convective
growth and decay are only partly predictable. Deterministic video predictors minimize a per-pixel
regression loss and are therefore driven toward the conditional mean of the future,
so their forecasts blur~\cite{convlstm,trajgru,predrnn,simvp,earthformer}. The blur erases
sharp, high-intensity cores, and detection skill at the heaviest
thresholds decays toward zero as the lead time grows. This failure mode is shared
even by strong recent predictors such as PhyDNet~\cite{phydnet} and
AlphaPre~\cite{alphapre}, because a mean-seeking objective smooths extremes
regardless of the backbone. A faithful
nowcaster must therefore be generative, sampling sharp and plausible fields rather
than averaging them.

Existing generative nowcasters address sharpness but leave two gaps. Many recent
diffusion-based methods anchor the forecast on a deterministic backbone,
refining it or adding a stochastic residual to it~\cite{cascast,diffcast};
because that backbone is trained with a pixel loss, it carries conditional-mean
blur, which the anchored forecast inherits at long lead times. Purely generative models, including
GAN- and flow-based nowcasters~\cite{dgmr,flowcast}, avoid the anchor but discard the
predictable part of the dynamics, namely the advective motion that a simple
physical model already captures well, and must relearn it from data, which can drift
the forecast off the observed storm track. These gaps
motivate a design that retains the
motion skill of a physical model while letting a generative model render the
unpredictable detail, without either contaminating the other.
\Cref{fig:teaser} contrasts deterministic, purely generative, and our
physics-guided designs.

In this paper we propose \methodfull{} (\method{}), a generative nowcaster that
guides a flow-map generative head with a frozen physical motion model. A lightweight
Lagrangian advection prior is pre-trained once and then frozen; it rolls the observed
field forward through a learned velocity and source term and supplies an
explicit motion forecast. A flow-map generative head then produces the future
sequence conditioned on both the history and this prior rollout. A central design
choice is that the prior enters as a conditioning signal rather than as an additive
residual, so the head can replace the prior's blurred structure with sharper detail
while still exploiting its motion skill, and the physical rollout keeps the convection
on a plausible trajectory. \revised{Here \emph{physics-guided} means that the
prior's rollout is a discretized advection equation in which only the
coefficient fields are learned (\cref{sec:prior}).}
Casting the generator as a two-time flow map rather than an instantaneous
velocity field lets it produce a forecast in a few network evaluations, in line
with recent flow-map and consistency models~\cite{flowmapmatching,consistencymodels}. This separation
of predictable motion from stochastic detail follows the decoupling logic proposed
in~\cite{nagashima2026neural}.

Across four standard radar benchmarks and six metrics each, \method{} outperforms the
state-of-the-art AlphaPre~\cite{alphapre} on $18$ of the $24$ metrics,
and the margins are largest precisely at the heavy-rain thresholds where mean-seeking
methods fail. \revised{Code and pretrained checkpoints are publicly
available.}\footnote{\url{https://github.com/Neurogica/PG-FMM}} The main
contributions of this work are summarized as follows.
\begin{itemize}
  \item We propose \method{}, which guides a generative flow-map head with a frozen
  Lagrangian advection prior used as conditioning rather than an additive
  residual, so conditional-mean blur is never inherited additively.
  \item We introduce, to our knowledge, the first flow-map-matching formulation for
  precipitation nowcasting, coupling a learned semi-Lagrangian advection prior with a
  two-time transport head whose few-step sampling makes the inference-time ensemble
  affordable.
  \item We show on the SEVIR, MeteoNet, Shanghai, and CIKM benchmarks that
  \method{} surpasses AlphaPre on $18$ of $24$ metrics, with the largest gains at the
  heavy-rain thresholds, and we isolate the prior's contribution with controlled
  ablations, which show that its benefit stems from the physics content rather
  than from conditioning alone.
\end{itemize}

%% file: tex/sections/02_related_work.tex
\subsection{Video Prediction and Precipitation Nowcasting}

Radar nowcasting is a conditional video-prediction problem, and the two literatures
have co-evolved since ConvLSTM~\cite{convlstm}, which was introduced for nowcasting
and became a standard video-prediction baseline~\cite{survey_videopred,survey_ffs}.
Deterministic sequence predictors developed on video transfer directly to radar,
including trajectory-aware recurrence (TrajGRU~\cite{trajgru}, the PredRNN
family~\cite{predrnn,predrnnpp,mim}, MotionRNN~\cite{motionrnn}), efficient
convolutional designs (SimVP~\cite{simvp}, MAU~\cite{mau}), and transformer and
operator backbones (Earthformer~\cite{earthformer}, Rainformer~\cite{rainformer});
physically structured predictors such as PhyDNet~\cite{phydnet} and the recent
amplitude--phase model AlphaPre~\cite{alphapre}, the strongest deterministic
predictor on the benchmarks we use, continue this line, and surveys cover both the
video and the weather perspective~\cite{survey_nowcasting,survey_weather}. Because
a squared- or absolute-error objective is minimized by the conditional mean, all of
these models tend to blur, and the blur is most damaging at high reflectivity where
sharp convective cells are smoothed away.

Stochastic video generation addresses the blur by sampling futures rather than
averaging them, as in video-diffusion predictors such as MCVD~\cite{mcvd} and
STRPM~\cite{strpm}. On radar, DGMR~\cite{dgmr} uses a conditional GAN, and
NowcastNet~\cite{nowcastnet} couples a deterministic evolution operator with a
generative network. Diffusion and latent-diffusion models have since become
dominant. PreDiff~\cite{prediff} and LDCast~\cite{ldcast} perform latent
diffusion, CasCast~\cite{cascast} cascades a deterministic prediction with a
latent diffusion refiner, and DiffCast~\cite{diffcast} adds a stochastic residual
on top of an arbitrary deterministic backbone. We refer to the latter recipe as a
residual anchor: the forecast is a deterministic prediction plus a generated
residual. Because the deterministic part is trained with a pixel loss, it carries
conditional-mean blur that the sum cannot remove, particularly at long lead times.
Diffusion models for the broader weather-forecasting
problem~\cite{gencast,precipdownscaling,omnicast} share the same generative
motivation at coarser scales.

\subsection{Flow Matching and Flow Maps}

Continuous-time generative models have matured rapidly, and recent
surveys~\cite{survey_diffusion,survey_flowmatching} cover the progression from
score-based diffusion~\cite{ddpm,scoresde,edm} to simulation-free transport. Flow
Matching~\cite{flowmatching}, Rectified Flow~\cite{rectifiedflow}, and Stochastic
Interpolants~\cite{stochasticinterpolants} regress an instantaneous velocity field
whose probability-flow ODE must be integrated with many steps at sampling time.
Flow-Map Matching~\cite{flowmapmatching} instead learns the finite-time flow map
(the two-time transport operator), so that a sample is produced in a few network
evaluations, or even one; Consistency Models~\cite{consistencymodels}, Consistency
Trajectory Models~\cite{ctm}, and Shortcut Models~\cite{shortcutmodels} reach the
same few-step regime through self-consistency, trajectory matching, and step-size
conditioning, and distillation offers a complementary route~\cite{progressivedistillation};
data-to-data diffusion bridges~\cite{i2sb,ddbm} are related transport
formulations. In nowcasting, FlowCast~\cite{flowcast}
applies conditional flow matching in a latent space. We adopt a flow-map head because
its few-step sampling makes a multi-member inference ensemble affordable, which is
otherwise prohibitive for long diffusion chains.

\subsection{Physics-Informed and Lagrangian Forecasting}

Physics-informed machine learning injects physical knowledge into data-driven models
through soft PDE-residual penalties, structured operators, or explicit dynamics, and
is reviewed comprehensively by Karniadakis~\etal~\cite{survey_piml,pinn}. In
precipitation nowcasting, a long operational tradition extrapolates the radar field
along estimated motion, as in the optical-flow libraries PySTEPS~\cite{pysteps} and
rainymotion~\cite{rainymotion}; such Lagrangian advection is accurate for short
horizons but cannot represent growth or decay. Learning-based work reintroduces the
missing source term while keeping the motion explicit: NowcastNet~\cite{nowcastnet}
learns a neural evolution operator with an additive intensity update, PhyDNet~\cite{phydnet}
disentangles physical dynamics from residual factors in a latent space, and the
differentiable Lagrangian network LUPIN~\cite{lupin} separates advection from an
advection-free growth/decay branch. Beyond nowcasting, large data-driven weather
models~\cite{fourcastnet,pangu,graphcast,neuralgcm} extend physically informed
architectures to global forecasting. Our prior follows the explicit
warp-and-source route but keeps the physics module lightweight and,
after a short pre-training stage, frozen, so that it supplies motion skill as a
conditioning signal, not a constraint on the generative head.

%% file: tex/sections/03_preliminaries.tex
\subsection{Problem Setting}

We address radar precipitation nowcasting. A model observes a history of $\Tin$ past
radar frames and predicts the next $\Tout$ frames, where each frame is a
single-channel reflectivity field $R \in \Reals^{H\times W}$ on a spatial grid of
size $H\times W$ normalized to $[0,1]$. We write the observed history as
$R_{1:\Tin}$ and the prediction target as
\begin{equation}
x_0 = R_{\Tin+1:\,\Tin+\Tout}\in\Reals^{\Tout\times H\times W},
\label{eq:target}
\end{equation}
where the $\Tout$ future frames are stacked along the channel axis for the
convolutional networks. The horizon lengths $\Tin,\Tout$, the grid size $H\times W$,
and the per-dataset intensity scale used to recover physical units are
dataset-specific and are listed in \cref{sec:exp}.

Because convective growth and decay are only partly determined by the recent past,
the future is inherently uncertain. We therefore model the conditional distribution
$p(x_0\mid R_{1:\Tin})$ and draw an ensemble of forecasts from it rather than
predicting a single field. A useful forecast combines high detection skill at the
operational intensity thresholds, especially the heavy-rain thresholds, with low
pixel error and sharp structure. We report the Critical Success Index
(CSI) per threshold and averaged, the Heidke Skill Score (HSS), SSIM, and MSE, with
formal definitions deferred to \cref{sec:exp}; the primary metrics are the mean CSI
and the high-threshold CSI.

\subsection{Flow Matching and Flow Maps}

Continuous-time generative models transport a simple base density to the data
distribution along an interpolant between a noise sample $x_1\sim\mathcal{N}(0,I)$ and
a data sample $x_0$,
\begin{equation}
x_\tau = (1-\tau)\,x_0 + \tau\,x_1,
\label{eq:prelim-interp}
\end{equation}
where $\tau\in[0,1]$ indexes the path from data ($\tau{=}0$) to noise ($\tau{=}1$).
Flow matching learns an instantaneous velocity field $v_\theta(x_\tau,\tau)$ whose
probability-flow ODE carries the base density to the data density, and sampling
integrates this ODE, which typically requires many function evaluations.

Flow-Map Matching instead learns the two-time solution operator of the same ODE. The
flow map $\flowmap_{t\to r}$ takes a state at time $t$ directly to the corresponding
state at an earlier time $r$,
\begin{equation}
\flowmap_{s\to r}\circ\flowmap_{t\to s}=\flowmap_{t\to r},
\label{eq:semigroup}
\end{equation}
where $0\le r<s<t\le 1$ and the semigroup identity holds because a flow map is a
solution operator. Enforcing this composition consistency during training regularizes
the learned operator, and because the map spans a finite time interval a sample is
produced in a few jumps rather than many small steps. Our generative head
(\cref{sec:method}) instantiates a conditional flow map of this form.

%% file: figures/overview.tex
%
%
\begin{figure}[t]
  \centering
  \includegraphics[width=\linewidth]{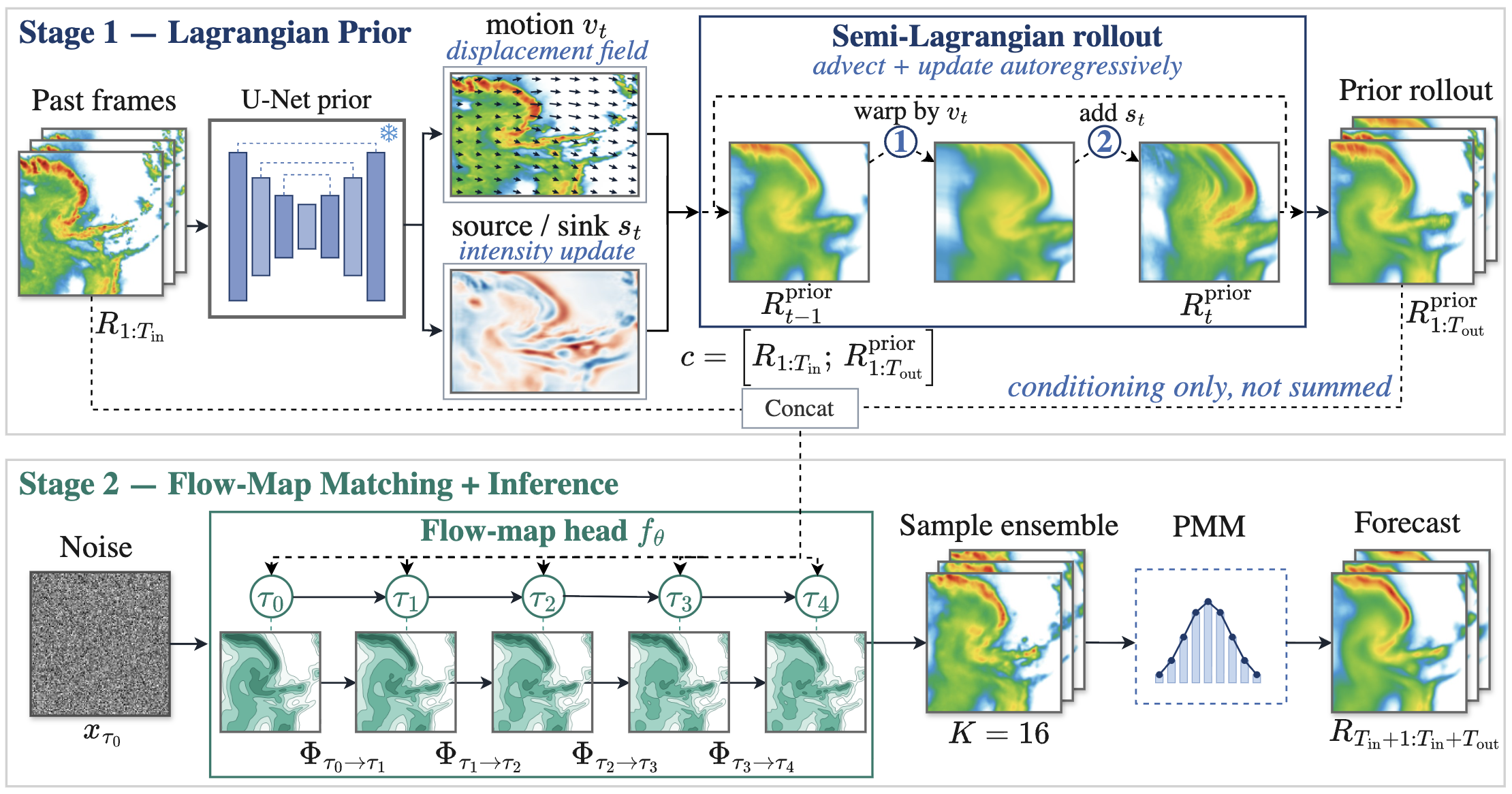}
  \caption{Overview of \method{}. A frozen Lagrangian advection prior $g_\phi$
  produces an explicit motion forecast, and a flow-map head $f_\theta$, conditioned
  on the history and the prior rollout, maps noise to the future sequence in a few
  transport steps. Stage 1 is pre-trained and frozen, Stage 2 is trained with the
  flow-map objective, and PMM denotes the probability-matched mean over $K$
  samples. The prior is never added to the output.}
  \label{fig:overview}
\end{figure}

%% file: tex/sections/04_method.tex
\subsection{Overview}

\method{} forecasts precipitation by separating the predictable advective motion of
the radar field from the uncertain small-scale detail, so that neither part
contaminates the other. The model has two modules (\cref{fig:overview}): a frozen
Lagrangian advection prior $g_\phi$ (\cref{sec:prior}) that turns the observed
history into an explicit motion forecast, and a flow-map generative head
$f_\theta$ (\cref{sec:flowmap}) that, conditioned on the history and the prior
rollout, samples the future sequence. At inference we aggregate a small ensemble
with a probability-matched mean (\cref{sec:inference}).

The prior is used only as conditioning and is never added to the output, unlike
residual-anchor nowcasters~\cite{cascast,diffcast}, whose pixel-loss backbone
carries its conditional-mean blur into the anchored forecast (\cref{sec:related}).
Passing the physical forecast as a conditioning signal instead lets the
generative head replace the prior's blurred structure with sharp detail while
still exploiting its motion skill. The decoupling does not depend on the
particular modules, as any per-step motion model and any conditional generative
head could be substituted. Ours pairs a learned warp-and-source prior with a
flow-map head whose few-step sampling keeps the ensemble affordable.

\subsection{Lagrangian Advection Prior}
\label{sec:prior}

Precipitation evolves partly through predictable transport and partly through
stochastic growth and decay, and a lightweight physical model can already account for the
first part. The prior captures this predictable motion explicitly, so the generative
head models only the residual uncertainty rather than relearning bulk advection
from scratch. The prior is implemented as a U-Net $g_\phi$ that ingests the stacked history
$R_{1:\Tin}$ and emits,
for each future step $t$, a dense velocity field $v_t\in\Reals^{2\times H\times W}$
and a scalar source/sink field $s_t\in\Reals^{H\times W}$. These two fields are
the only learned quantities of Stage~1; the rollout operator below contains no
parameters. Both outputs are bounded for stability as
\begin{equation}
v_t = d_{\max}\,\tanh(\tilde v_t),\qquad
s_t = c_s\,\tanh(\tilde s_t),
\label{eq:bounds}
\end{equation}
where $\tilde v_t,\tilde s_t$ are the raw network outputs, $d_{\max}$ caps the
per-step displacement, and $c_s$ caps the intensity update. Starting from the
last observed frame, the forecast is rolled out autoregressively as
\begin{equation}
\Rp_t = \clipop_{[0,1]}\!\Big(\warpop\big(\Rp_{t-1}, v_t\big) + s_t\Big),
\quad t=1,\dots,\Tout,
\quad \Rp_0 = R_{\Tin},
\label{eq:lagrangian}
\end{equation}
where $\warpop(\cdot,v)$ is backward bilinear sampling and $\clipop_{[0,1]}$ keeps
intensities in the normalized range. The update is a semi-Lagrangian
discretization of the advection equation with a source term,
$\partial_t R + v\cdot\nabla R = s$. The warp evaluates each output pixel at its
upstream location, displaced backward along $v_t$, which is the Lagrangian view
of tracking precipitation parcels along the flow, so existing intensity is transported across the grid
rather than re-synthesized; the additive source $s_t$ models the growth and
decay that pure advection cannot represent. \revised{Learning is thus confined
to the coefficient fields $v_t$ and $s_t$. The rollout operator itself has no
parameters, so every forecast of the prior satisfies the discretized advection
equation by construction.}

Stage~1 is supervised regression rather than generative modeling. The prior is
trained alone, before the generative head, to forecast the observed future
frames under the composite objective
\begin{equation}
\mathcal{L}_{\text{prior}} =
\ell_1^{w} + \lambda_{\text{mse}}\,\mathrm{MSE}
+ \lambda_{\text{csi}}\,\mathcal{L}_{\text{soft-CSI}}
+ \lambda_{v}\,\lVert\nabla v\rVert_1 + \lambda_{s}\,\lVert s\rVert_1,
\label{eq:priorloss}
\end{equation}
where $\ell_1^{w}=\Expect\big[(1{+}2R)\,\lvert\hat R-R\rvert\big]$ is an
intensity-weighted $\ell_1$ distance between the rollout $\hat R{=}\Rp_t$ and the
observed future frame $R$ (mean over pixels, frames, and the batch),
$\mathcal{L}_{\text{soft-CSI}}$ is a differentiable sigmoid relaxation of the CSI
detection metric of \cref{sec:exp} (construction in the supplementary material),
$\lVert\nabla v\rVert_1$ and $\lVert s\rVert_1$ penalize rough motion fields and
overactive sources so that advection explains as much of the change as possible,
and $\lambda_{\text{mse}}$, $\lambda_{\text{csi}}$, $\lambda_{v}$, and
$\lambda_{s}$ are scalar weights (values in the supplementary material).
After this stage the prior is frozen and receives no gradient during generative
training or at inference.

\subsection{Flow-Map Matching Generative Head}
\label{sec:flowmap}

A deterministic prior cannot supply sharp, calibrated detail, so the second
module is generative. It models the conditional distribution of the future frames given the
history and the prior rollout, and it is realized as a flow map rather than an
instantaneous velocity field, so that sampling needs only a few network
evaluations and the inference ensemble (\cref{sec:inference}) remains affordable. We treat
the future sequence $x_0$ as the data endpoint of a linear stochastic
interpolant~\cite{stochasticinterpolants} between Gaussian noise
$x_1\sim\mathcal{N}(0,I)$ and data, as in \cref{eq:prelim-interp}. The conditioning
$c$ concatenates, along the channel axis, the observed history $R_{1:\Tin}$ and the
frozen prior rollout $\Rp_{1:\Tout}$ from \cref{eq:lagrangian}; it is appended to the
network input at every evaluation and, again, never added to the output.

Instead of regressing the instantaneous velocity of the interpolant, the network
learns its flow map: $f_\theta(x_t, c; t, r)$ predicts the state $x_r$ directly from
the state $x_t$ for $0\le r<t\le1$~\cite{flowmapmatching}; here $t$ and $r$ are
interpolant times, not frame indices. The two times enter through a timestep embedding and as two constant channel
planes. The backbone is a U-Net with self-attention at its
coarsest scale; its width and depth are reported in the supplementary material. The primary loss
regresses the operator to the true interpolant endpoint as
\begin{equation}
\mathcal{L}_{\text{FM}} =
\Expect_{x_0,x_1,(t,r)}\big\lVert f_\theta(x_t, c; t, r) - x_r \big\rVert_2^2 ,
\label{eq:fm}
\end{equation}
where the times satisfy a minimum gap $t-r\ge\delta_{\min}$ and, with probability
$p_{0}$, we force $r{=}0$ so the network also learns the direct noise-to-data map used
at the first sampling step. A single large jump regresses toward a conditional
average; multi-step sampling (\cref{sec:inference}) and the consistency term
below counteract this. So that the learned map behaves as a transport
operator, we add a composition-consistency regularizer that enforces the
semigroup identity of \cref{eq:semigroup}, writing
$\flowmap_{t\to r}(\cdot)\equiv f_\theta(\cdot, c;t,r)$ with the conditioning $c$
shared by every factor. For a random intermediate time $s\in(r,t)$, composing $t\to s$
and then $s\to r$ must reproduce the direct map $t\to r$; violations are
penalized as
\begin{equation}
\begin{split}
\mathcal{L}_{\text{CC}} = {}&
\underbrace{\tfrac12\lVert\flowmap_{t\to s}(x_t)-x_s\rVert_2^2
+\tfrac12\lVert\flowmap_{s\to r}(\flowmap_{t\to s}(x_t))-x_r\rVert_2^2}_{\text{sub-map fidelity}} \\
&+\underbrace{\lVert\flowmap_{s\to r}(\flowmap_{t\to s}(x_t))-\flowmap_{t\to r}(x_t)\rVert_2^2}_{\text{self-consistency}},
\end{split}
\label{eq:cc}
\end{equation}
where $x_s$ is the interpolant state at the intermediate time $s$
(\cref{eq:prelim-interp}), the expectation is over $x_0,x_1$ and times $r<s<t$
drawn as in \cref{eq:fm}, and the intermediate prediction is detached before the
second map; the two fidelity terms coincide with \cref{eq:fm} evaluated at
$(t,s)$ and $(s,r)$. The total
objective is $\mathcal{L}_{\text{FM}}+\lambda_{\text{CC}}\,\mathcal{L}_{\text{CC}}$,
with the schedule for $\lambda_{\text{CC}}$ given in the supplementary material; in our setting
the consistency term stabilizes training and improves fidelity.

\subsection{Inference: Few-Step Sampling and Ensemble}
\label{sec:inference}

Because $f_\theta$ is a transport operator, a forecast is produced with a small number
of function evaluations (NFE). On a decreasing time grid
$1{=}\tau_0>\tau_1>\dots>\tau_{N}{=}0$, we initialize $x_{\tau_0}\sim\mathcal{N}(0,I)$
and iterate as
\begin{equation}
x_{\tau_{i+1}} = f_\theta\big(x_{\tau_i}, c; \tau_i, \tau_{i+1}\big),
\qquad i=0,\dots,N-1,
\label{eq:sampling}
\end{equation}
returning $\clipop_{[0,1]}(x_{\tau_{N}})$, where $N$ is the number of sampling
steps; each step is one evaluation of $f_\theta$, so a forecast costs exactly
$N$ NFE. We use exponential-moving-average weights of the trained network. A
single draw is sharp, but its small-scale detail varies with the noise seed. Few-step sampling makes it inexpensive to draw an ensemble of $K$ independent
samples from different noise seeds and combine them with a probability-matched
mean (PMM)~\cite{pmm}. Averaging the members directly would blur again, since
displaced peaks cancel; the PMM instead takes the spatial pattern of the
ensemble mean, which places precipitation reliably, and remaps its values, rank
by rank, onto the value distribution pooled over all $K$ members, so the
high-amplitude peaks that plain averaging smooths away are restored, which is
decisive for high-threshold CSI. We report
the full PMM for verification metrics and, for qualitative figures, a single sample,
which is visually sharper; the operating-point values $N$ and $K$ are given in
\cref{sec:exp}.

%% file: tex/sections/05_experiments.tex
\subsection{Datasets and Protocol}

We evaluate on four standard radar benchmarks spanning different regions and
sensors: SEVIR (United States; vertically integrated liquid,
VIL)~\cite{sevir}, MeteoNet (France)~\cite{meteonet}, Shanghai Radar
(China)~\cite{shanghai}, and CIKM~\cite{cikm}. Following the AlphaPre
protocol~\cite{alphapre}, every frame is resized to
$H{\times}W{=}128{\times}128$ and normalized to $[0,1]$, and we predict $\Tout$
future frames from $\Tin{=}5$ past frames, with $\Tout{=}20$ on SEVIR, MeteoNet
and Shanghai and $\Tout{=}10$ on CIKM. We use each benchmark's official splits
throughout: SEVIR provides $12{,}182$ training and $8{,}096$ test sequences;
MeteoNet $6{,}308$ / $1{,}310$; Shanghai $1{,}381$ / $526$ (a $10\%$ tail of the
official $1{,}534$ training sequences is held out for validation, with no test
leakage); and CIKM $1{,}000$ / $4{,}000$. Each dataset keeps its native
intensity scale, thresholds and evaluator, so all numbers are comparable.

\subsection{Implementation Details}

\revised{The two modules are trained in sequence with AdamW (learning rate $10^{-4}$,
batch size $8$) and exponential-moving-average weights (decay $0.999$). The
Lagrangian prior is trained first and then frozen; the flow-map head is then
trained for $297.6$k steps for the headline SEVIR model, and every ablation
variant uses a matched $150$k-step budget. The prior U-Net uses base width $96$ with
channel multipliers $(1,2,4,4)$; the flow-map U-Net uses base width $128$
with multipliers $(1,2,3,4)$ and self-attention at the two coarsest scales.
At inference we use $N{=}4$ sampling steps and a $K{=}16$-member
probability-matched-mean ensemble, i.e., $64$ network evaluations per
forecast ($1.2$ forecasts/s on one GPU; a single sample runs at $7.2$/s).
Training and evaluation run on one NVIDIA GB202 GPU ($96$\,GB), and checkpoints
are selected by validation skill.} Full hyperparameters, per-dataset training
steps, and timing are listed in the supplementary material.

\subsection{Baselines and Metrics}
\input{tex/tables/tab_main}
\begin{figure}[t]
  \centering
  \includegraphics[width=0.78\linewidth]{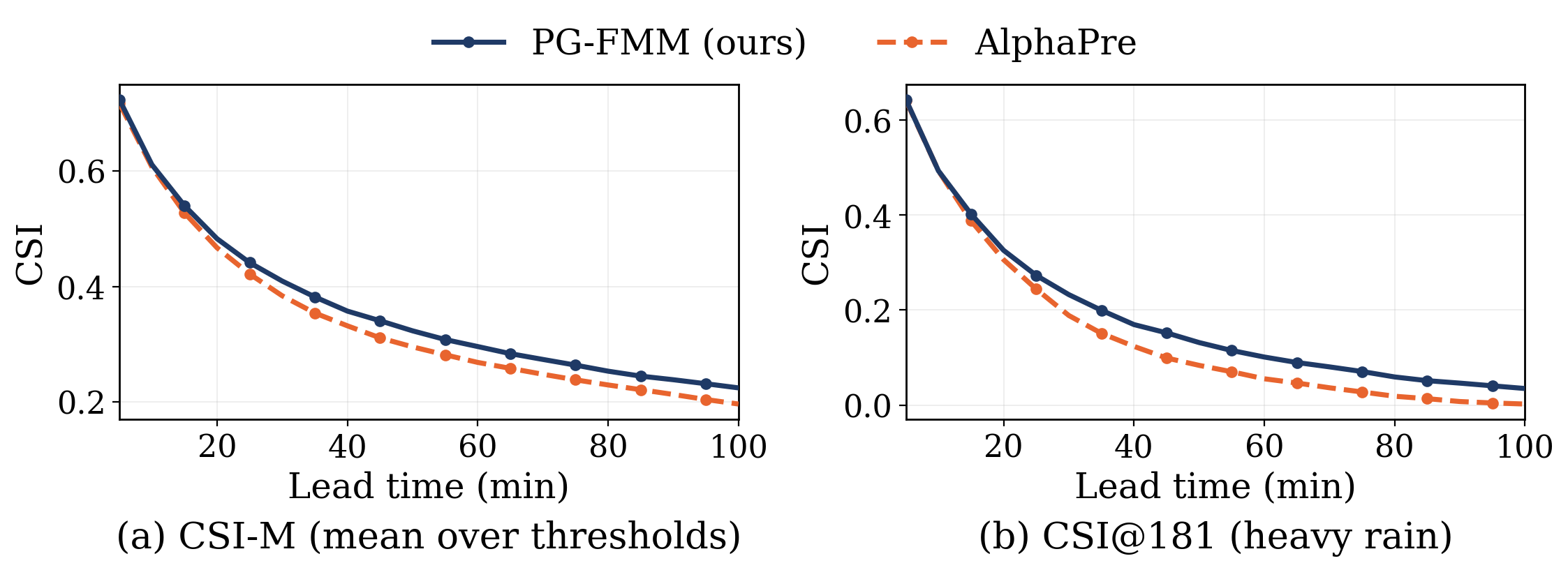}
  \caption{Detection skill versus lead time on SEVIR: CSI averaged over
  thresholds (a) and at the heavy-rain threshold VIL${=}181$ (b). The margin of
  \method{} over AlphaPre widens with lead time and is largest on heavy rain.}
  \label{fig:leadtime}
\end{figure}

We compare against the baselines reported in the AlphaPre study under the same
protocol. These include the deterministic predictors ConvGRU~\cite{trajgru},
MAU~\cite{mau}, SimVP~\cite{simvp}, FourCastNet~\cite{fourcastnet},
Earthformer~\cite{earthformer}, PhyDNet~\cite{phydnet}, and
Earthfarseer~\cite{earthfarseer}, and the strongest recent deterministic and
generative methods, NowcastNet~\cite{nowcastnet}, DiffCast~\cite{diffcast}, and
AlphaPre~\cite{alphapre}.

We report the standard meteorological detection and fidelity metrics. For a
binarization threshold $\kappa$, counting hits ($\mathrm{TP}$), misses ($\mathrm{FN}$),
and false alarms ($\mathrm{FP}$) over the thresholded prediction and ground truth, the
Critical Success Index and Heidke Skill Score are
\begin{equation}
\mathrm{CSI}_\kappa=\tfrac{\mathrm{TP}}{\mathrm{TP}+\mathrm{FN}+\mathrm{FP}},\quad
\mathrm{HSS}=\tfrac{2(\mathrm{TP}\,\mathrm{TN}-\mathrm{FP}\,\mathrm{FN})}
{(\mathrm{TP}{+}\mathrm{FN})(\mathrm{FN}{+}\mathrm{TN})+(\mathrm{TP}{+}\mathrm{FP})(\mathrm{FP}{+}\mathrm{TN})} ,
\label{eq:csi}
\end{equation}
where $\mathrm{TN}$ counts correct negatives. CSI-M is the mean CSI over the
dataset's thresholds, and we additionally report CSI at the two highest
(heavy-rain) thresholds, the hardest regime. We also report SSIM and MSE for
pixel fidelity, following the benchmark protocol; SSIM rewards smoothness, so
we treat it as a reference value, and the primary metrics are CSI-M and the
high-threshold CSI. As our model predicts a distribution, we additionally
report the continuous ranked probability score (CRPS)~\cite{gneiting2007}, a
strictly proper scoring rule computed from the $16$-member ensemble, for our
model and its generative ablation variants, which share one operating point;
the \cref{tab:main} baselines are transcribed from the AlphaPre benchmark,
which defines no CRPS protocol.

\subsection{Main Results}

\begin{figure}[t]
  \centering
  \includegraphics[width=0.70\linewidth]{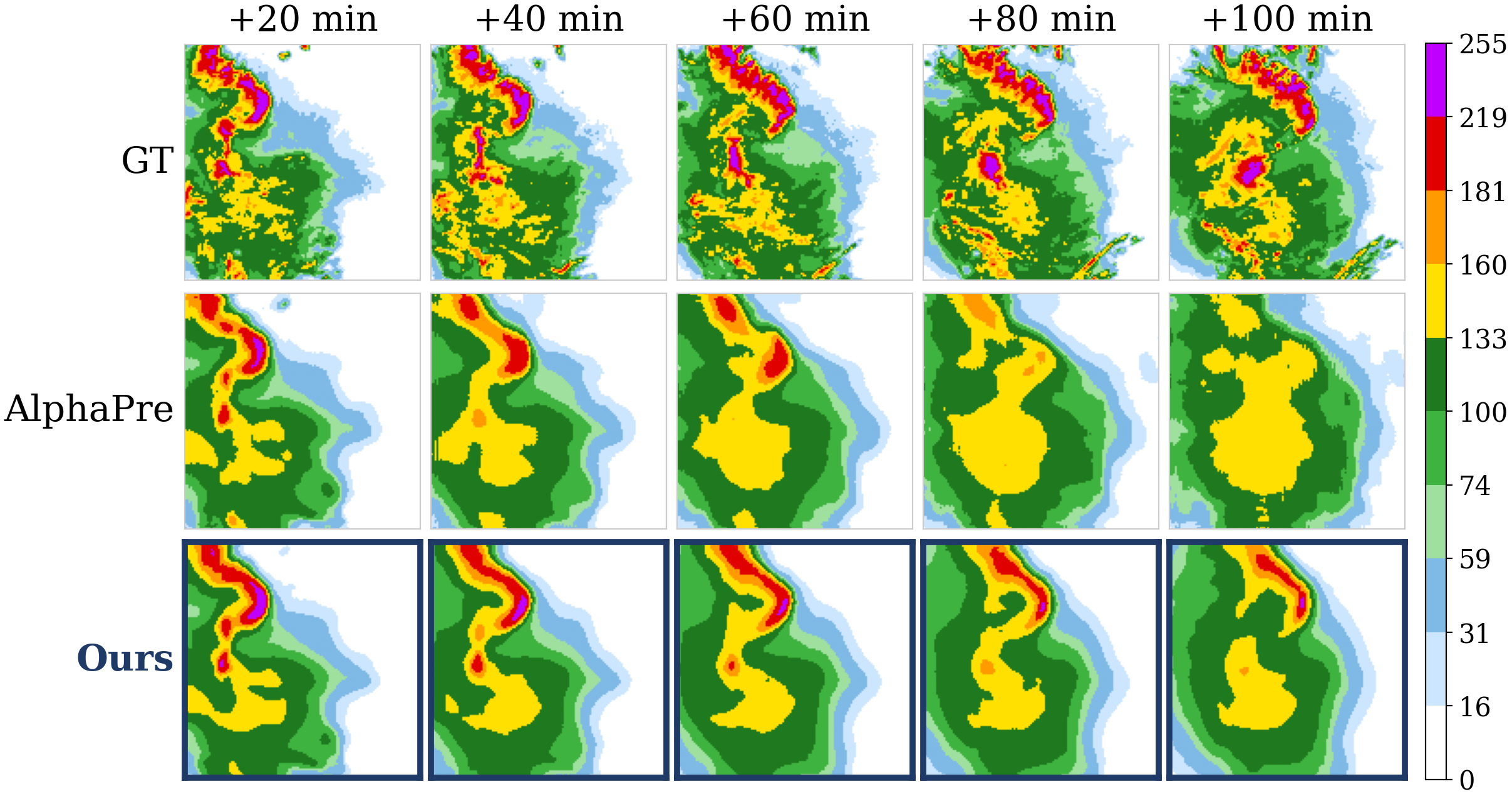}\\[4pt]
  \includegraphics[width=0.70\linewidth]{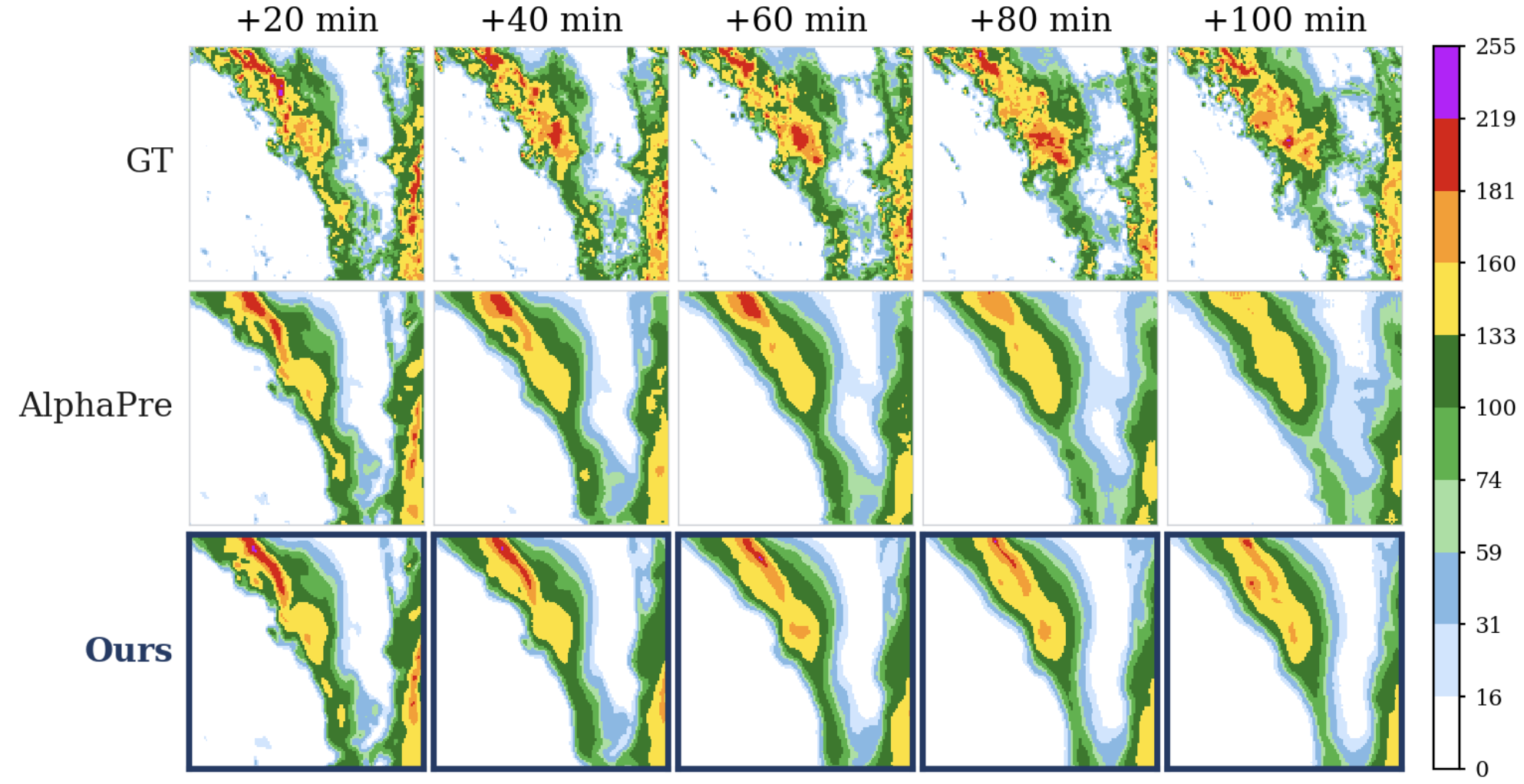}
  \caption{Qualitative forecasts on SEVIR-VIL (rows: ground truth, AlphaPre,
  \method{}; columns: $+20$ to $+100$ minutes). Top: a convective system over
  Nebraska (test index $1708$). Bottom: the remnants of Hurricane Barry (15 July 2019). \method{}
  preserves the heavy-rain cores and peak intensities that AlphaPre blurs.}
  \label{fig:qual}
\end{figure}

Across the four benchmarks and six metrics each, \method{} outperforms AlphaPre
on $18$ of the $24$ metrics (\cref{tab:main}). It is best on every metric on
SEVIR and MeteoNet, and the gains concentrate at the heavy-rain thresholds: on
the heaviest SEVIR threshold CSI rises from $0.0582$ to $0.0925$, a $58.9\%$
relative gain. \Cref{fig:leadtime} traces this margin over lead time, the CSI-M
gap widening from $+0.006$ at $+5$ minutes to $+0.028$ at $+100$ minutes while
the baseline's heavy-rain detection decays toward zero. On Shanghai, \method{}
is best on four of six metrics, including CSI-40, SSIM, MSE and, narrowly, mean
CSI; on CIKM it is best on the two heavy-rain thresholds and second on HSS.
\method{} also outperforms the strongest generative baseline, DiffCast, on every
SEVIR metric, and its $16$-member ensemble attains a CRPS of $0.0294$ on SEVIR,
with a spread--skill diagnostic in the supplementary material. Sampling-seed variability at the reported operating point is negligible, far
below the reported margins (supplementary material).

\input{tex/tables/tab_ablation}
\subsection{Qualitative Results}

\Cref{fig:qual} shows two cases over the full $100$-minute horizon
(single samples, the sharper qualitative view; verification uses the
probability-matched mean, \cref{sec:inference}). On the
Nebraska convective system (top), \method{} preserves the sharp, high-intensity
core at every lead time while AlphaPre progressively attenuates and blurs it;
the frozen prior keeps the convection on its observed track while the generative
head restores intensity. The same behavior holds on the remnants of Hurricane
Barry (bottom).

\begin{figure}[t]
  \centering
  \includegraphics[width=0.70\linewidth]{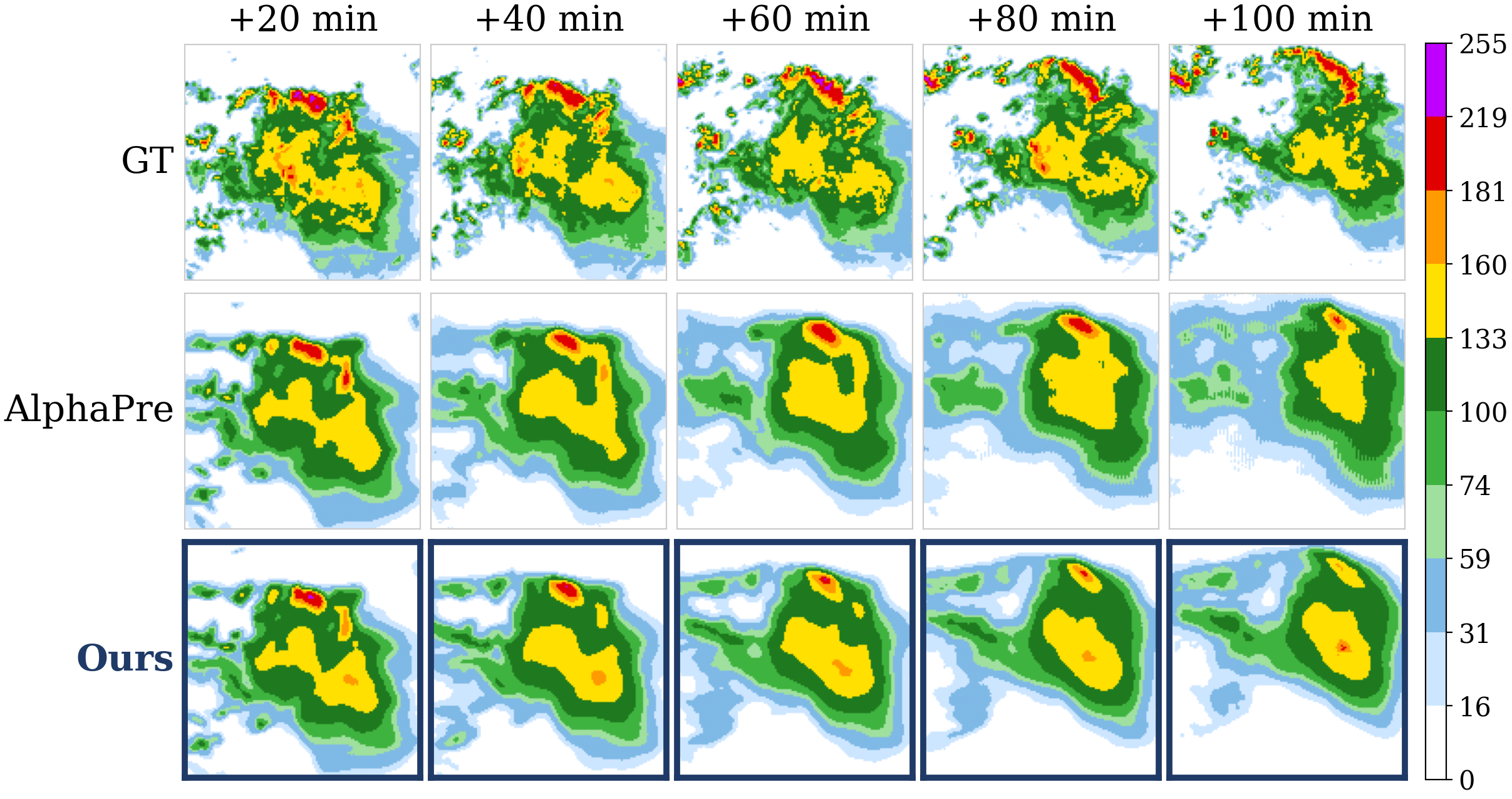}
  \caption{A representative failure case, a convective system over north Texas
  (test index $103$; rows and colors as in \cref{fig:qual}). With growing lead
  time \method{} drops the heavy band that both the ground truth and AlphaPre
  retain (CSI-160 $0.218$ vs.\ $0.299$).}
  \label{fig:failure}
\end{figure}

To test whether these examples are representative rather than favorable, we
ranked \emph{all} $2{,}672$ heavy-rain SEVIR test samples (ground-truth max VIL
$\geq 120$) by the per-sample CSI-160 gain over AlphaPre at the reported
operating point; AlphaPre attains the higher CSI-160 on $351$ of them
($13.1\%$). \Cref{fig:failure} shows a representative failure, a convective
system over north Texas where the generative head drops the heavy-rain band
that both the ground truth and the deterministic baseline retain. The
supplementary material provides additional cases on SEVIR and MeteoNet.

\subsection{Ablation Study}

Four generative variants share an identical flow-map head, identical
conditioning, and a matched $150$k-step training budget on SEVIR, and
differ only in the content of the prior rollout (\cref{tab:ablation}): no prior,
classical semi-Lagrangian advection driven by optical flow, a learned deterministic
nowcaster (AlphaPre) used as the prior, and our frozen Lagrangian physics prior. The
physics prior is best on every metric except SSIM, followed by the learned
deterministic prior, and classical advection trails even the no-prior variant
on every detection metric. The content of the prior is therefore what
matters: a poor prior is worse than no prior at all, so the gains cannot be
explained by conditioning alone. The one exception, SSIM, rewards smoother
output, as the Prior only row below shows most clearly. The full model in
\cref{tab:main} uses a longer schedule, hence its slightly higher numbers;
per-lead-time curves in the supplementary material show that the physics
prior's margin emerges at long lead times.

The generative head is likewise necessary. The frozen prior alone (Prior only in
\cref{tab:ablation}) is already a strong deterministic baseline, yet it trails
the full model on every skill metric; its best-in-table SSIM reflects
smoothness, and the head adds the sharpness and probabilistic skill (CRPS).
Together, these ablations justify the motion--generative decoupling.

\revised{The supplementary material reports further studies. Composition
consistency (\cref{eq:cc}) appears to improve fidelity and stability rather
than detection skill, so we keep it enabled. An ensemble-size sweep supports
the $16$-member operating point, and even a single draw ($K{=}1$, CSI-M
$0.3394$) at $N{=}4$ network evaluations exceeds AlphaPre's published CSI-M.
A fusion ablation indicates that conditioning, unlike additive residual
fusion, is robust to the quality of the prior. Structural metrics suggest
that the heavy-rain gain reflects recovered heavy rain rather than inflated
area, and a matched-ensemble comparison with a reproduced generative baseline
indicates that the margin persists.}
\FloatBarrier

%% file: tex/tables/tab_main.tex
\begin{table}[t]
  \centering
  \caption{Comparison on four radar benchmarks under the AlphaPre evaluation
  protocol~\cite{alphapre}; ours uses a $16$-member probability-matched mean.
  ND/D denotes models without/with an explicit dynamics module, and CSI-$x$ is
  the CSI at threshold $x$, the dataset's two heavy-rain levels. In each column,
  the best result is in \textbf{bold} and the second best is \underline{underlined}.}
  \label{tab:main}
  \setlength{\tabcolsep}{4pt}
  \setlength{\aboverulesep}{0pt}\setlength{\belowrulesep}{0pt}
  \renewcommand{\arraystretch}{1.12}
  \resizebox{\textwidth}{!}{%
  \begin{tabular}{l l | c c c c c c | c c c c c c}
    \toprule
    & & \multicolumn{6}{c|}{\textbf{SEVIR}} & \multicolumn{6}{c}{\textbf{MeteoNet}} \\
    \cmidrule(lr){3-8}\cmidrule(lr){9-14}
    Model & Type & CSI-M$\uparrow$ & CSI-181$\uparrow$ & CSI-219$\uparrow$ & HSS$\uparrow$ & SSIM$\uparrow$ & MSE$\downarrow$ & CSI-M$\uparrow$ & CSI-24$\uparrow$ & CSI-32$\uparrow$ & HSS$\uparrow$ & SSIM$\uparrow$ & MSE$\downarrow$ \\
    \midrule
    ConvGRU      & ND & 0.2903 & 0.0879 & 0.0350 & 0.3619 & 0.6100 & 368.34 & 0.3401 & 0.2990 & 0.1431 & 0.4667 & 0.7833 & 12.85 \\
    MAU          & ND & 0.3076 & 0.1071 & 0.0516 & 0.3863 & 0.6505 & 355.48 & 0.3233 & 0.2839 & 0.0997 & 0.4452 & 0.7897 & 12.92 \\
    SimVP        & ND & 0.3108 & 0.1106 & 0.0517 & 0.3924 & 0.6508 & 383.56 & 0.3351 & 0.3002 & 0.1130 & 0.4573 & 0.7804 & 13.45 \\
    FourCastNet  & ND & 0.2686 & 0.0717 & 0.0339 & 0.3355 & 0.5976 & 410.27 & 0.3027 & 0.2533 & 0.1085 & 0.4216 & 0.6450 & 15.05 \\
    Earthformer  & ND & 0.2892 & 0.0844 & 0.0245 & 0.3665 & 0.6633 & 360.11 & 0.3205 & 0.2884 & 0.1237 & 0.4491 & 0.7772 & 14.43 \\
    PhyDNet      & D  & 0.3017 & 0.1040 & 0.0278 & 0.3812 & 0.6532 & 357.63 & 0.3384 & 0.3194 & 0.1366 & 0.4673 & 0.7823 & 14.48 \\
    Earthfarseer & D  & 0.3004 & 0.0992 & 0.0413 & 0.3829 & 0.6327 & 388.91 & 0.3404 & 0.3170 & 0.1372 & 0.4726 & 0.7542 & 14.10 \\
    NowcastNet   & D  & 0.2791 & 0.0770 & 0.0351 & 0.3512 & 0.6839 & 412.94 & 0.3427 & 0.3206 & 0.1598 & 0.4751 & 0.7879 & 15.64 \\
    DiffCast     & D  & 0.3050 & 0.1300 & \underline{0.0582} & 0.3996 & 0.6482 & 559.59 & 0.3512 & 0.3340 & 0.1808 & 0.4846 & 0.7887 & 17.93 \\
    AlphaPre     & D  & \underline{0.3259} & \underline{0.1332} & 0.0545 & \underline{0.4110} & \underline{0.6884} & \underline{345.18} & \underline{0.3824} & \underline{0.3633} & \underline{0.2002} & \underline{0.5164} & \underline{0.7968} & \underline{12.74} \\
    \rowcolor{blue!8}
    \textbf{Ours} & D & \textbf{0.3614} & \textbf{0.1859} & \textbf{0.0925} & \textbf{0.4593} & \textbf{0.7291} & \textbf{315.67} & \textbf{0.4239} & \textbf{0.4066} & \textbf{0.2293} & \textbf{0.5583} & \textbf{0.8426} & \textbf{9.25} \\
    \midrule
    & & \multicolumn{6}{c|}{\textbf{Shanghai}} & \multicolumn{6}{c}{\textbf{CIKM}} \\
    \cmidrule(lr){3-8}\cmidrule(lr){9-14}
    Model & Type & CSI-M$\uparrow$ & CSI-35$\uparrow$ & CSI-40$\uparrow$ & HSS$\uparrow$ & SSIM$\uparrow$ & MSE$\downarrow$ & CSI-M$\uparrow$ & CSI-35$\uparrow$ & CSI-40$\uparrow$ & HSS$\uparrow$ & SSIM$\uparrow$ & MSE$\downarrow$ \\
    \midrule
    ConvGRU      & ND & 0.3612 & 0.3163 & 0.2062 & 0.4899 & 0.7796 & 33.56 & 0.3091 & 0.2009 & 0.1259 & 0.4006 & 0.6507 & 37.13 \\
    MAU          & ND & 0.3983 & 0.3621 & 0.2417 & 0.5346 & 0.7195 & 30.40 & 0.3039 & 0.2054 & 0.1241 & 0.3928 & 0.6325 & 40.74 \\
    SimVP        & ND & 0.3850 & 0.3549 & 0.2382 & 0.5194 & 0.7795 & 34.40 & 0.3052 & 0.2044 & 0.1321 & 0.3955 & 0.6538 & 38.06 \\
    FourCastNet  & ND & 0.3571 & 0.3108 & 0.2073 & 0.4868 & 0.5598 & 32.10 & 0.2980 & 0.1849 & 0.1015 & 0.3801 & 0.4359 & \underline{36.14} \\
    Earthformer  & ND & 0.3503 & 0.3178 & 0.1872 & 0.4844 & 0.7298 & 35.57 & 0.3077 & 0.2039 & 0.1369 & 0.4001 & 0.6267 & 36.49 \\
    PhyDNet      & D  & 0.3654 & 0.3236 & 0.2176 & 0.4957 & 0.7751 & 36.41 & 0.3038 & 0.2052 & 0.1287 & 0.3931 & 0.6541 & 39.56 \\
    Earthfarseer & D  & 0.3926 & 0.3608 & 0.2343 & 0.5330 & 0.5405 & 32.68 & 0.3000 & 0.2046 & 0.1259 & 0.3911 & 0.6373 & 39.87 \\
    NowcastNet   & D  & 0.3953 & 0.3608 & 0.2450 & 0.5334 & 0.7902 & 33.56 & 0.2991 & 0.1940 & 0.1188 & 0.3865 & \textbf{0.6713} & 40.96 \\
    DiffCast     & D  & 0.4089 & 0.3740 & 0.2606 & 0.5476 & 0.7879 & 36.35 & \underline{0.3159} & 0.2009 & \underline{0.1457} & 0.4085 & 0.6499 & 42.78 \\
    AlphaPre     & D  & \underline{0.4178} & \textbf{0.3854} & \underline{0.2615} & \textbf{0.5534} & \underline{0.7951} & \underline{28.02} & \textbf{0.3194} & \underline{0.2068} & 0.1416 & \textbf{0.4137} & \underline{0.6568} & \textbf{35.18} \\
    \rowcolor{blue!8}
    \textbf{Ours} & D & \textbf{0.4181} & \underline{0.3819} & \textbf{0.2636} & \underline{0.5520} & \textbf{0.7962} & \textbf{26.05} & 0.3149 & \textbf{0.2218} & \textbf{0.1540} & \underline{0.4115} & 0.6252 & 42.18 \\
    \bottomrule
  \end{tabular}%
  }
\end{table}

%% file: tex/tables/tab_ablation.tex
\begin{table}[t]
  \centering
  \caption{Prior ablation on SEVIR ($16$-member PMM, four sampling steps), rows
  sorted by CSI-M; the generative variants differ only in the prior content,
  and Prior only additionally removes the generative head; we omit its CRPS,
  which degenerates to the pixel MAE for a deterministic forecast and is not
  comparable to ensemble CRPS. In each column, the best result is in
  \textbf{bold} and the second best is \underline{underlined}.}
  \label{tab:ablation}
  \setlength{\tabcolsep}{4pt}
  \setlength{\aboverulesep}{0pt}\setlength{\belowrulesep}{0pt}
  \renewcommand{\arraystretch}{1.12}
  \resizebox{\linewidth}{!}{%
  \begin{tabular}{l ccc cc cc}
    \toprule
    Prior & CSI-M$\uparrow$ & CSI-181$\uparrow$ & CSI-219$\uparrow$ & HSS$\uparrow$ & SSIM$\uparrow$ & MSE$\downarrow$ & CRPS$\downarrow$ \\
    \midrule
    Classical advection              & 0.3373 & 0.1490 & 0.0745 & 0.4260 & 0.7113 & 327.13 & 0.0308 \\
    None (pure generative)           & 0.3525 & 0.1690 & 0.0884 & 0.4466 & 0.6968 & 319.60 & 0.0312 \\
    Prior only (no generative head)  & 0.3552 & 0.1737 & 0.0859 & 0.4493 & \textbf{0.7330} & 334.95 & -- \\
    Deterministic (AlphaPre)         & \underline{0.3571} & \underline{0.1782} & \underline{0.0892} & \underline{0.4530} & \underline{0.7244} & \underline{317.87} & \underline{0.0299} \\
    \rowcolor{blue!8}
    Lagrangian physics (ours)        & \textbf{0.3601} & \textbf{0.1807} & \textbf{0.0907} & \textbf{0.4573} & 0.7226 & \textbf{315.28} & \textbf{0.0298} \\
    \bottomrule
  \end{tabular}%
  }
\end{table}

%% file: tex/sections/06_conclusion.tex
We addressed radar precipitation nowcasting in the high-intensity regime, where
existing models blur the strongest convective cells. \method{} decouples
predictable advection from uncertain small-scale detail and conditions a
flow-map head on a frozen physical prior rather than summing a residual onto
it, so no conditional-mean blur is additively inherited. Under a unified
protocol over four radar benchmarks, \method{} surpassed the state-of-the-art
AlphaPre on $18$ of $24$ metrics, with the largest margins at the heavy-rain
thresholds where mean-seeking methods fail, and controlled ablations isolated
the contributions of the prior and the head.

\revised{Two limitations remain. The raw ensemble is under-dispersive, with a
spread-to-RMSE ratio of about $0.43$ and a U-shaped rank histogram
(supplementary material). The probability-matched mean compensates for this
only at the verification level, so probabilistic use beyond the reported CRPS
and Brier scores requires explicit spread calibration. All results also come
from single-seed training on $128{\times}128$ rasters under one protocol with
lead times up to $100$ minutes, and, as with any advection-based nowcaster,
\method{} under-forecasts initiation and re-intensification without
precursors, can under-carry heavy bands (\cref{sec:exp}) or over-persist
dissipating storms (supplementary material). Addressing these limitations and
extending the decoupling to other spatiotemporal tasks are natural next steps.}

%% file: tex/sections/supp_body.tex
\section*{A\quad Skill versus Lead Time for the Prior Ablation}

\begin{figure}[!htb]
  \centering
  \includegraphics[width=\linewidth]{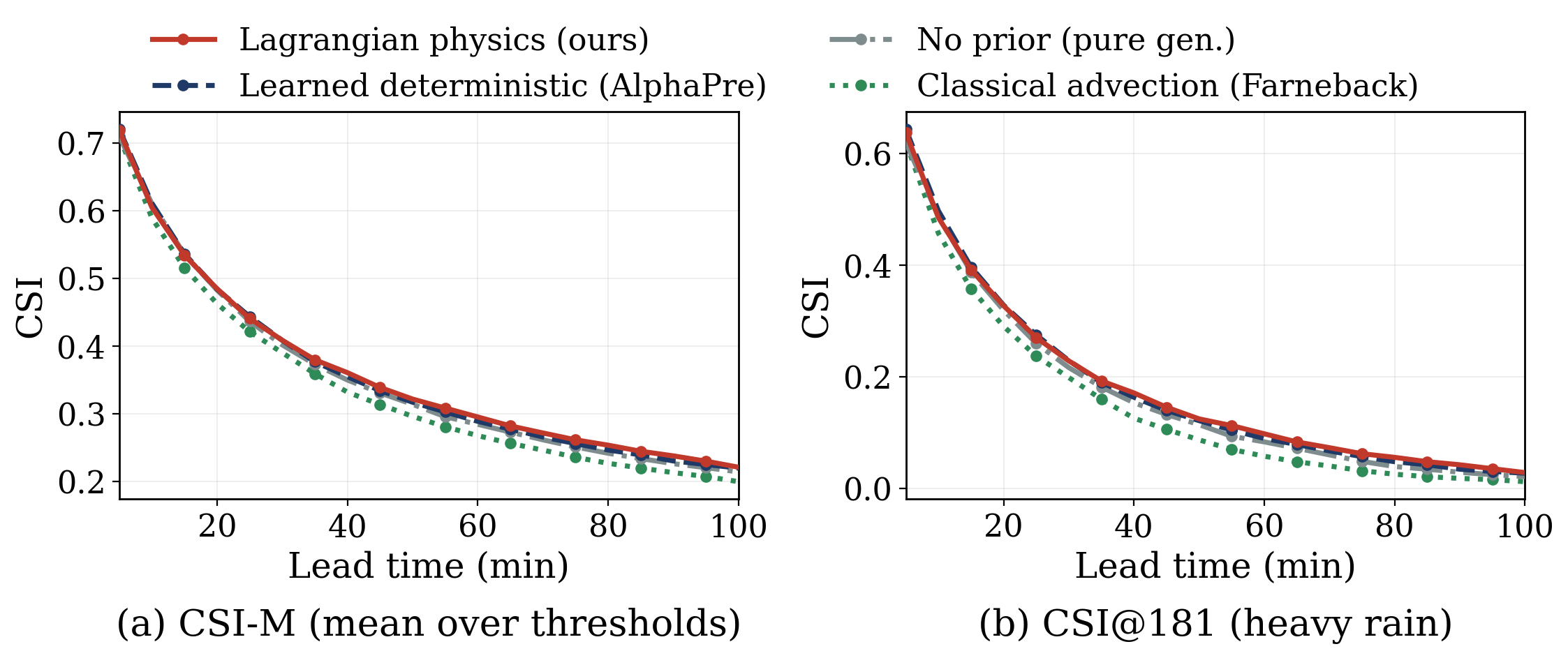}
  \caption{Skill versus lead time for the prior ablation on SEVIR: CSI averaged
  over thresholds (a) and at the heavy-rain threshold VIL${=}181$ (b). The margin of
  the physics prior emerges at long lead times and is largest on heavy rain.}
  \label{fig:leadtime_abl}
\end{figure}

Figure~\ref{fig:leadtime_abl} complements the prior ablation (Table~2 of the main paper)
with per-lead-time detection skill. All four variants share an identical flow-map
head, identical conditioning, and a matched training budget on SEVIR, and are
evaluated at the unified operating point ($16$-member probability-matched mean, four
sampling steps); only the content of the prior rollout differs.

Two observations mirror the aggregate table. First, the physics prior is
indistinguishable from the alternatives at short lead times, where advection is
easy, and its margin appears and grows at longer lead times, the regime where an
informative motion prior matters. Second, the classical advection prior degrades
fastest, falling below the no-prior variant, which indicates that a poor prior
actively misleads the generative head and that the gains of our prior come from
its content rather than from conditioning alone.

\section*{B\quad Spread--Skill Diagnostic and Sampling-Seed Variability}

\begin{figure}[!htb]
  \centering
  \includegraphics[width=0.95\linewidth]{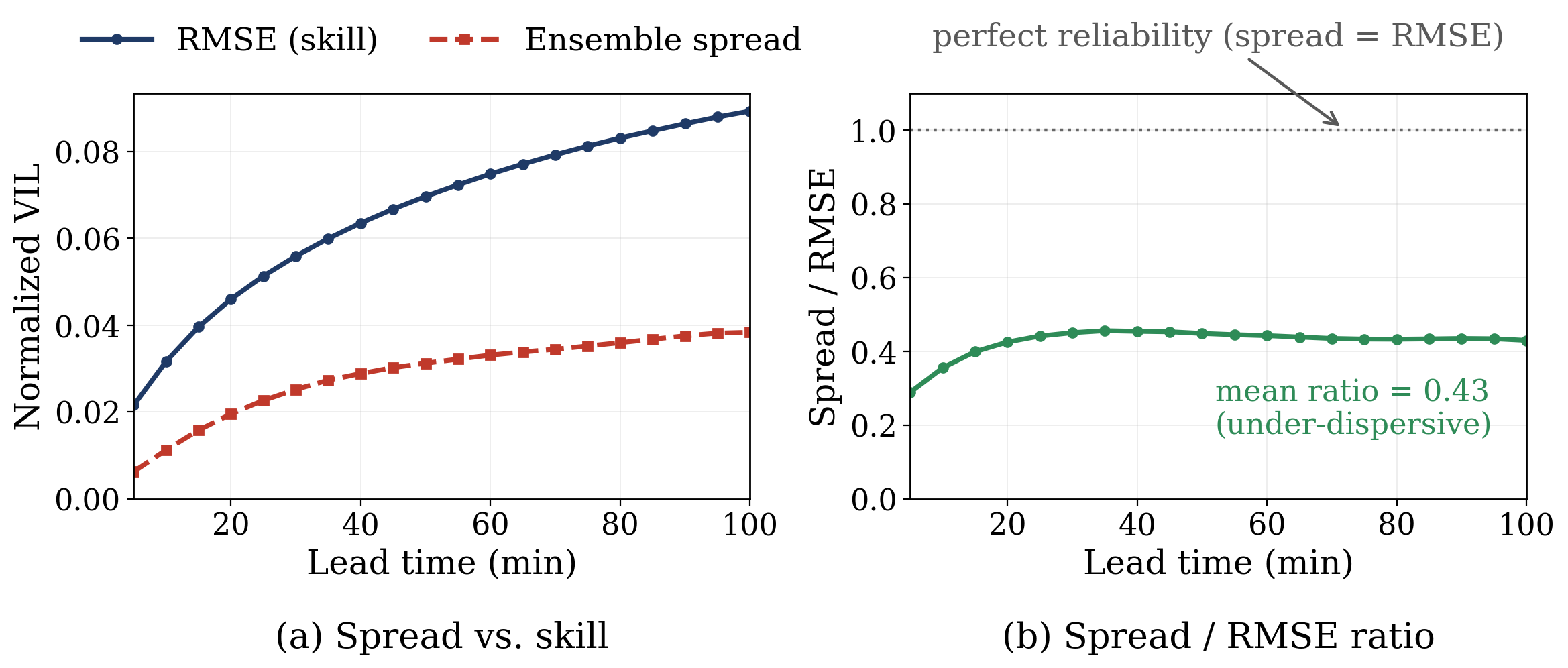}
  \caption{Spread--skill diagnostic of \mbox{PG-FMM} on SEVIR: ensemble spread
  and ensemble-mean RMSE per lead time (normalized VIL units). The spread/RMSE
  ratio is stable at ${\approx}0.43$, indicating a sharp but under-dispersive
  ensemble.}
  \label{fig:spreadskill}
\end{figure}

Figure~\ref{fig:spreadskill} reports the spread--skill relation of the full model
on the SEVIR test set, namely the ensemble spread (standard deviation across the
$16$ members) and the ensemble-mean RMSE per lead time, in normalized VIL units.
The spread grows monotonically with lead time, as the ensemble widens when the
forecast becomes less certain, and the spread-to-RMSE ratio is stable at
${\approx}0.43$. The ensemble is
therefore under-dispersive, the common regime for radar nowcast ensembles, which
trade calibration of the raw members for sharpness; this is consistent with the
strong CRPS of the probability-matched mean. Improving member-level dispersion, for
example through spread calibration or larger ensembles, is left as future work.

Sampling-seed variability at the reported operating point, summarized in the
main paper, is negligible. Across three independent sampling seeds, the CSI-M
standard deviation is $6{\times}10^{-5}$
and the heavy-rain CSIs vary by $2{\times}10^{-4}$ (the $16$-member
probability-matched mean averages out per-sample noise), one to two orders of
magnitude below every margin discussed in the main paper.

\section*{C\quad Ensemble-Size and Sampling-Step Sweeps}

\input{tex/tables/tab_inference}
\input{tex/tables/tab_nfe}

Table~\ref{tab:inference} complements the ablation study of the main paper with a
sweep over the ensemble size used by the probability-matched mean, on the same
SEVIR checkpoint and at four sampling steps. Every metric improves monotonically
with the number of members, but the returns diminish sharply. The CSI-M gain per
doubling roughly halves at each step, and moving from $16$ to $64$ members costs
$4\times$ the inference compute for $+0.0013$ CSI-M. We therefore use $16$ members
throughout the main paper as the accuracy/compute trade-off; few-step sampling
makes an ensemble of this size affordable at inference.

Table~\ref{tab:nfe} adds the corresponding sweep over the number of sampling
steps $N$ at a fixed $16$-member probability-matched mean. One step, the
setting most prone to conditional averaging, is the weakest on detection, and
multi-step sampling recovers monotonically; $N{=}4$ is the accuracy--compute
operating point used throughout.

\section*{D\quad Learned Velocity and Source Fields}

\begin{figure}[!htb]
  \centering
  \includegraphics[width=\linewidth]{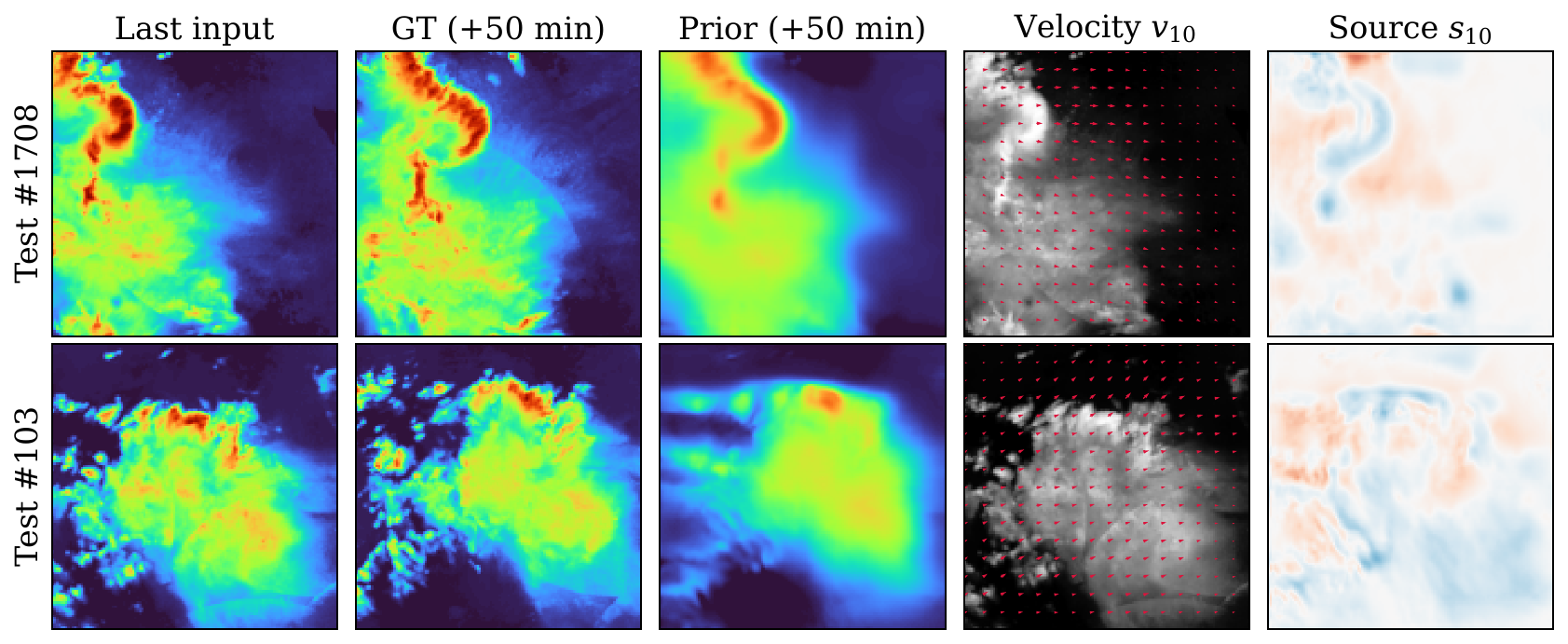}
  \caption{Learned fields of the frozen Lagrangian prior for two test cases
  (rows). Columns: last observed frame, ground truth and prior rollout at
  $+50$ minutes, the learned velocity $v_{10}$ (arrows, over the last frame),
  and the learned source $s_{10}$ (diverging colors, $\pm 0.05$ of the
  normalized intensity). The velocity forms a coherent advective flow, and the
  source is small and sparse (mean $|s|$ is $1.3\%$ of its bound $c_s{=}0.25$;
  $5.6\%$ of pixels exceed $0.01$), consistent with its role as a growth/decay
  correction. Over rain pixels the first-step velocity agrees with
  Farneb\"ack optical flow at a cosine similarity of $0.865$.}
  \label{fig:prior_fields}
\end{figure}

\section*{E\quad Probabilistic Diagnostics}

\begin{figure}[!htb]
  \centering
  \includegraphics[width=\linewidth]{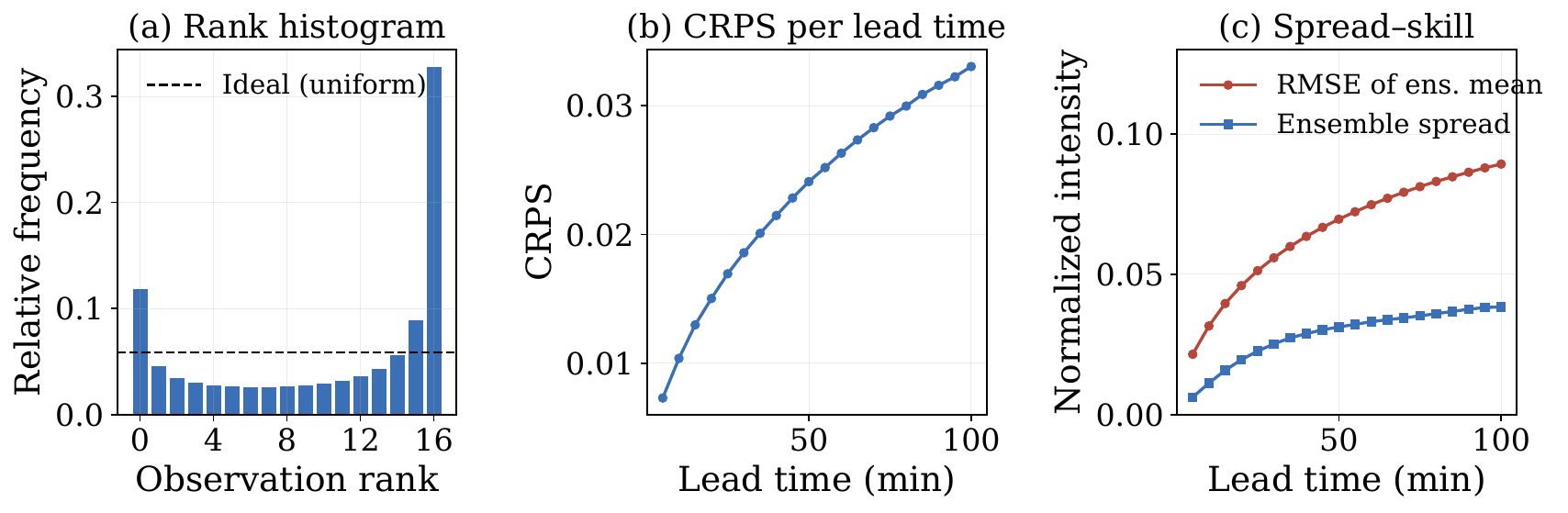}
  \caption{Probabilistic diagnostics of the $16$-member ensemble on the full
  SEVIR test set. (a)~The rank histogram over rain pixels is U-shaped and
  right-heavy, i.e., the raw ensemble is under-dispersive, consistent with the
  limitation stated in the conclusion; the probability-matched mean compensates
  at the verification level. (b)~CRPS grows monotonically with lead time.
  (c)~Ensemble spread stays below the RMSE of the ensemble mean
  (spread/RMSE $\approx 0.43$). Brier scores at VIL${=}16$--$219$ are
  $0.0583$, $0.0361$, $0.0151$, $0.0058$, $0.0027$, $0.0005$.}
  \label{fig:probabilistic}
\end{figure}
A representative failure case, selected systematically rather than by hand, is
discussed in the main paper together with the qualitative results. Section~D
below adds further success and failure cases.

\section*{F\quad Additional Qualitative Results}

\begin{figure}[p]
  \centering
  \includegraphics[width=0.82\linewidth]{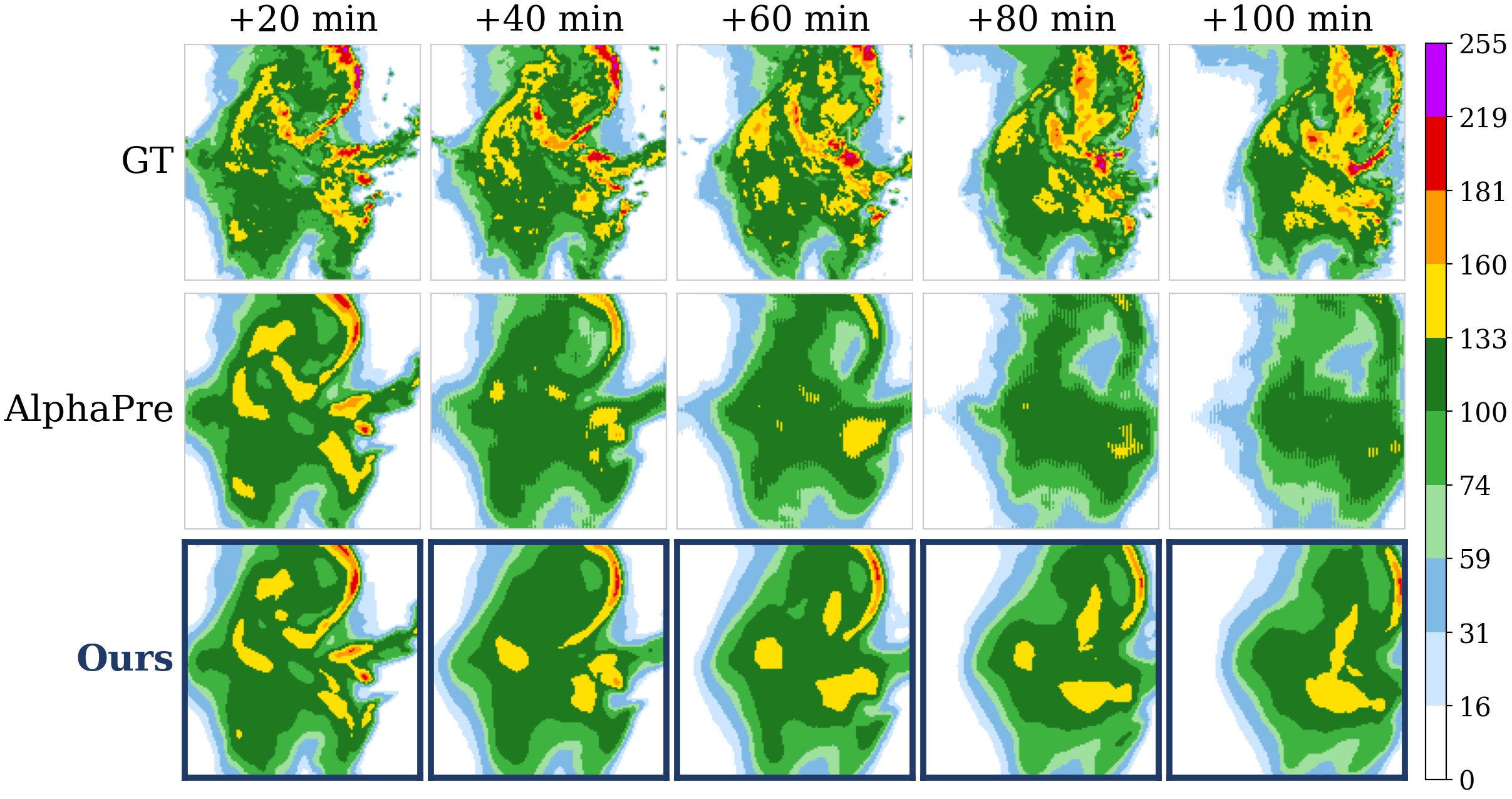}\\[4pt]
  \includegraphics[width=0.82\linewidth]{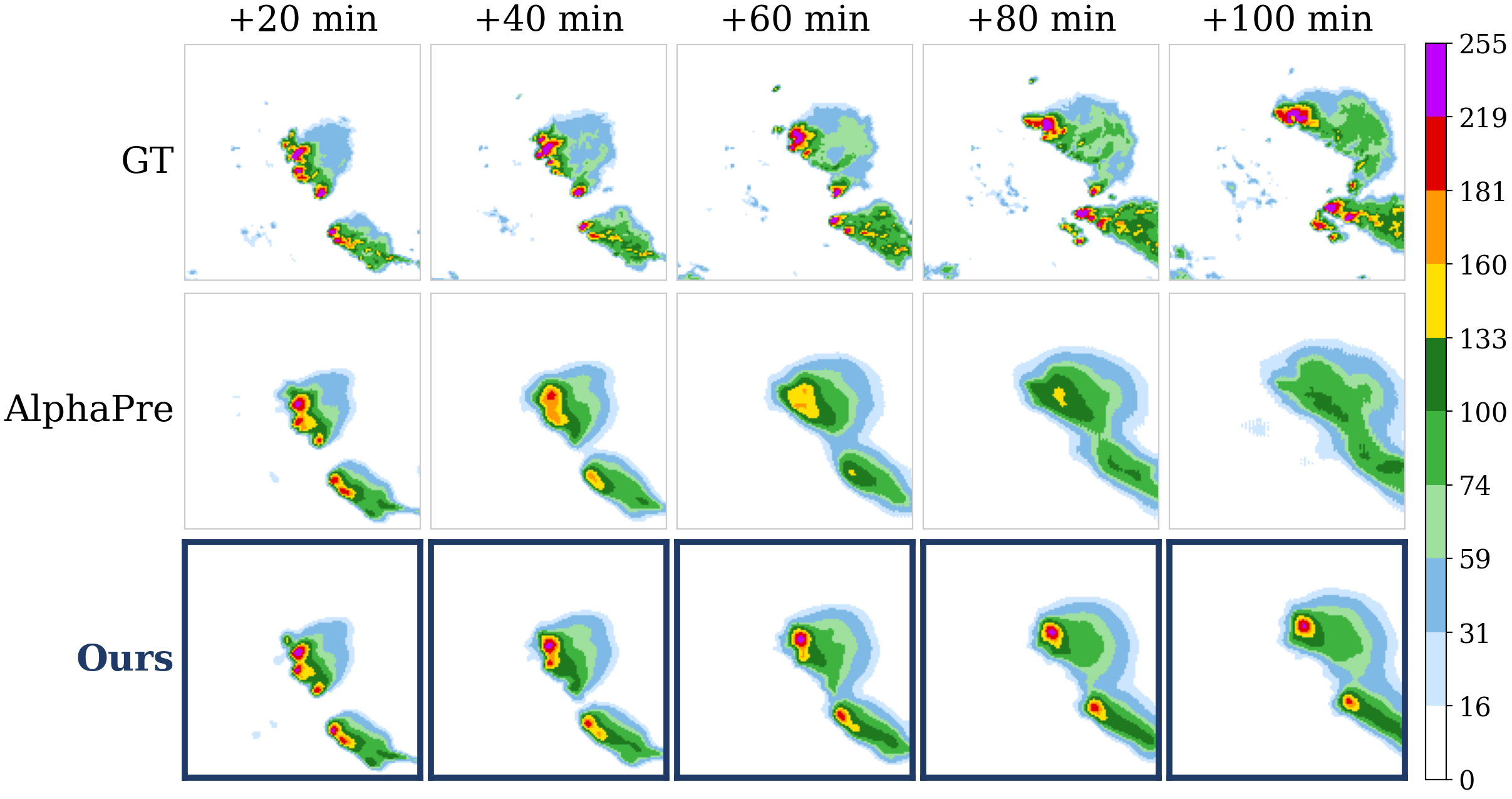}\\[4pt]
  \includegraphics[width=0.82\linewidth]{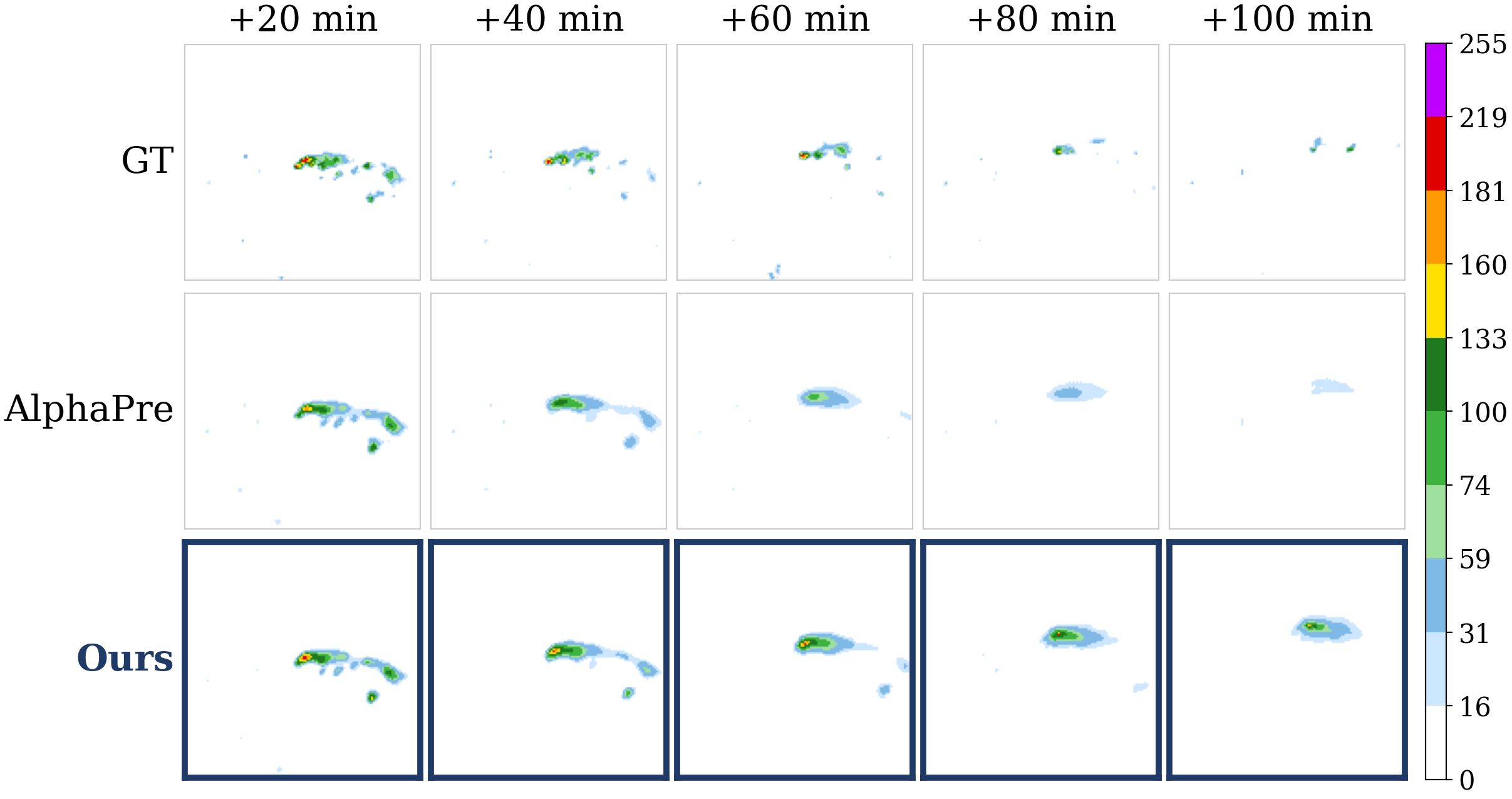}
  \caption{Additional SEVIR-VIL cases (rows: ground truth, AlphaPre, \mbox{PG-FMM};
  columns: $+20$ to $+100$ minutes). Top: a success on a frame-filling system
  (test index $1955$; CSI-160 $0.178$ vs.\ $0.147$). Middle: a success (index
  $7243$; $0.258$ vs.\ $0.221$). Bottom: a failure (index $635$; $0.196$ vs.\
  $0.211$), in
  which the observed storm dissipates by $+80$ minutes and AlphaPre correctly
  fades, whereas the generative head maintains a compact storm.}
  \label{fig:qual-supp-sevir}
\end{figure}

\begin{figure}[p]
  \centering
  \includegraphics[width=0.82\linewidth]{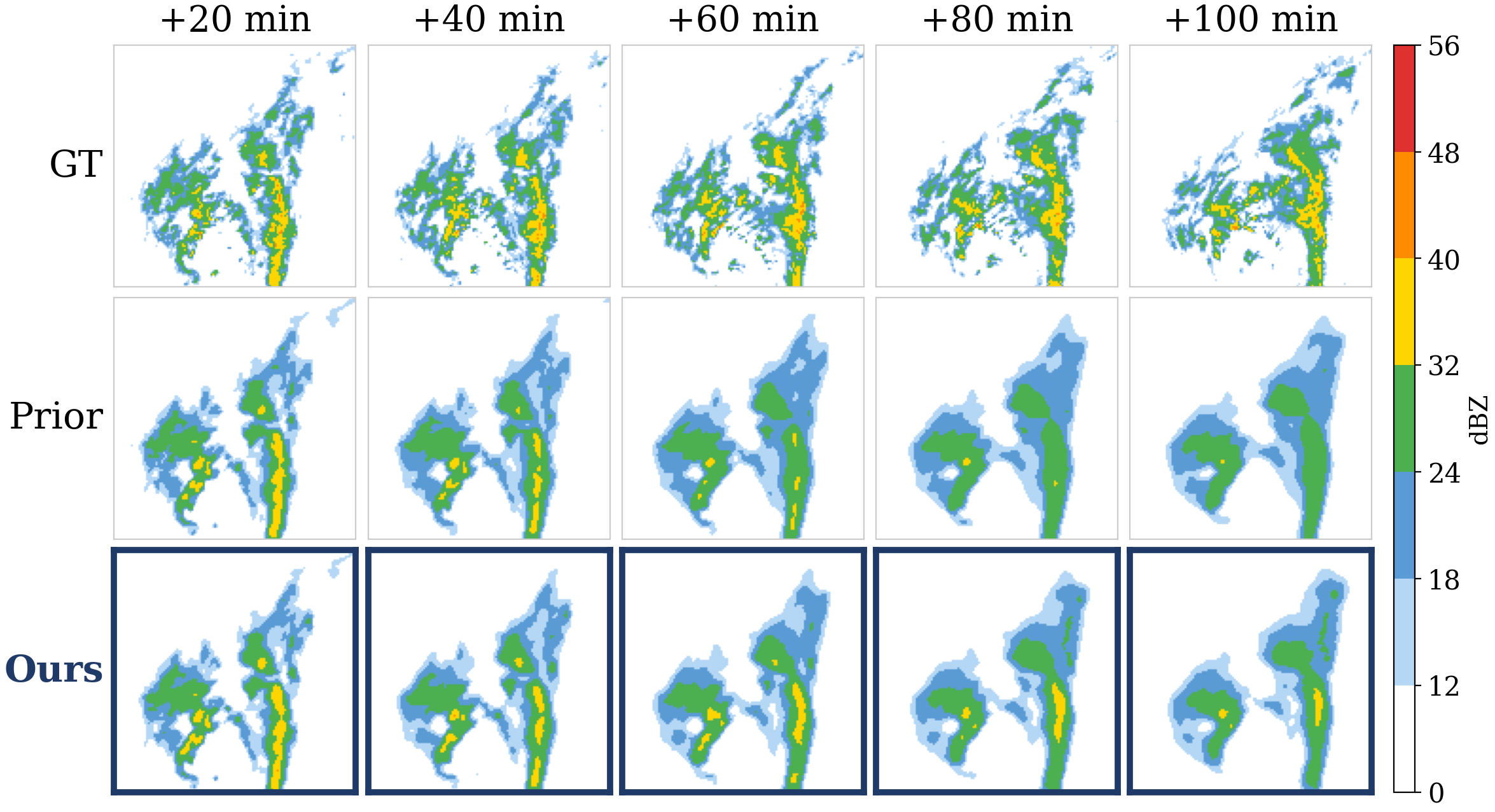}\\[4pt]
  \includegraphics[width=0.82\linewidth]{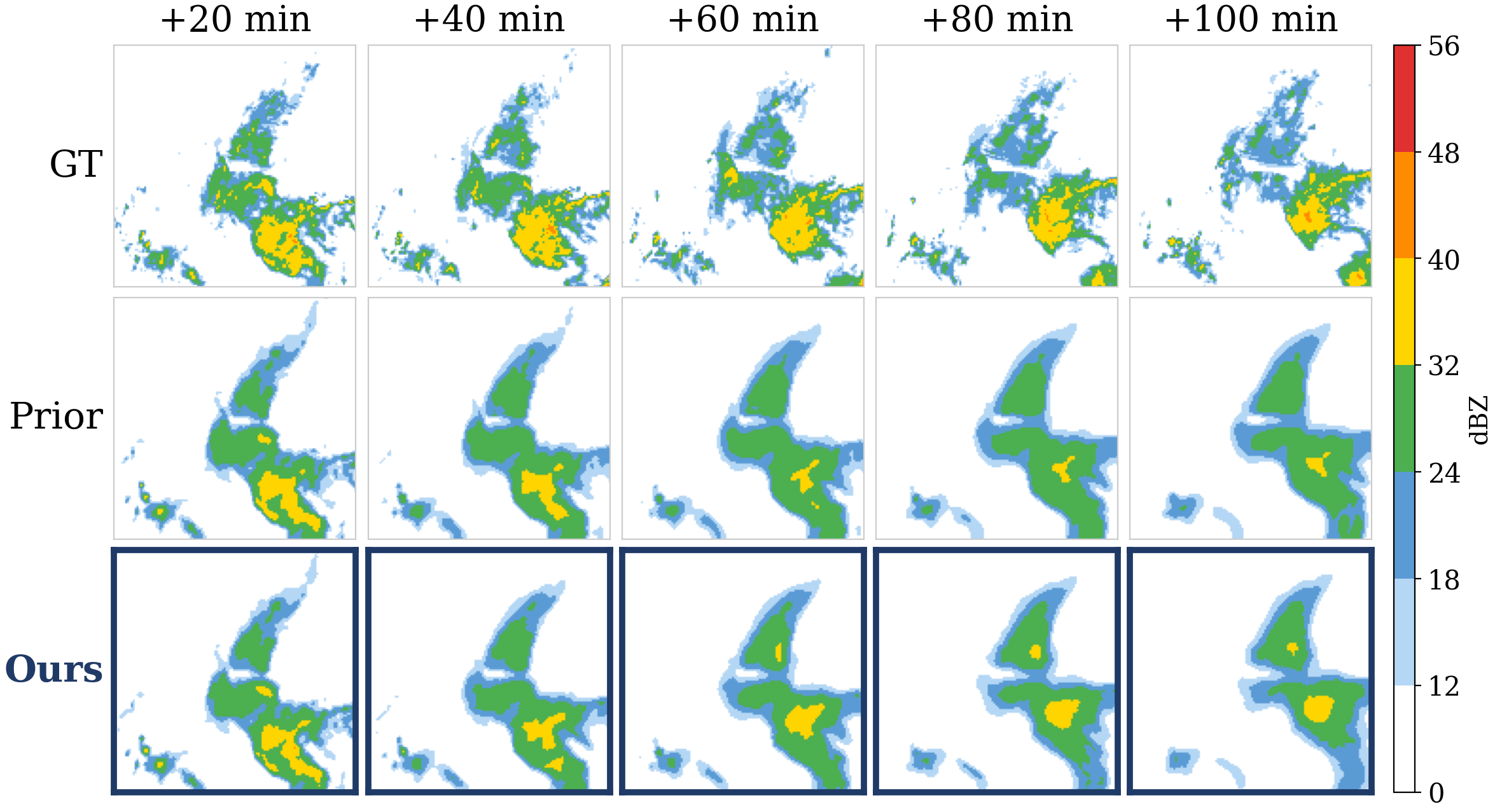}\\[4pt]
  \includegraphics[width=0.82\linewidth]{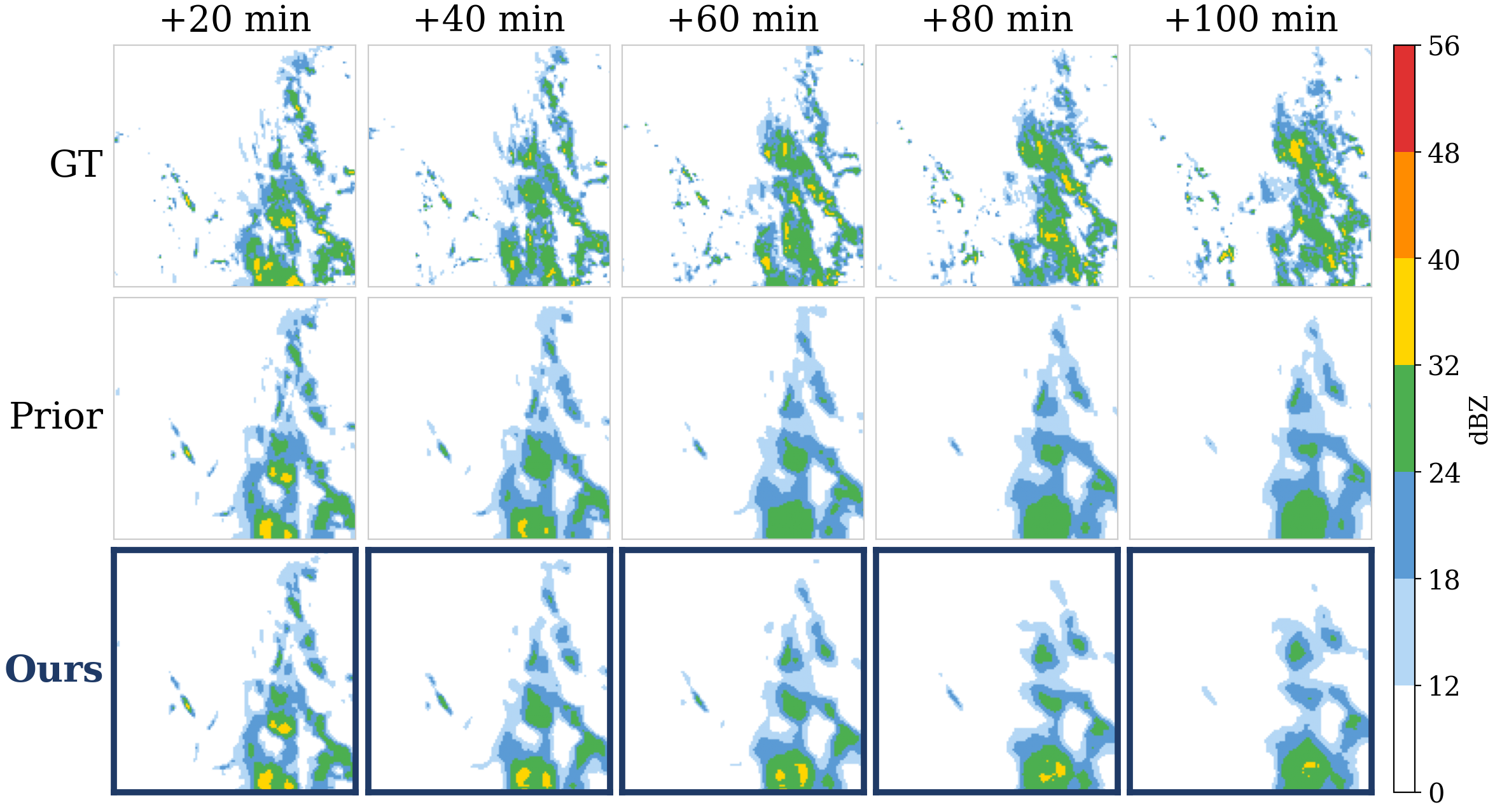}
  \caption{MeteoNet cases (rows: ground truth, the frozen advection prior,
  \mbox{PG-FMM}; columns: $+20$ to $+100$ minutes; reflectivity in dBZ). Top and
  middle: successes (test indices $1138$ and $962$), where the prior smooths the
  embedded $32$--$40$\,dBZ cells away with lead time and \mbox{PG-FMM} restores
  them. Bottom: a failure (index $460$), where the observed convection keeps
  regenerating heavy cells that the prior never carries and the head only
  partially restores.}
  \label{fig:qual-supp-meteo}
\end{figure}

Figures~\ref{fig:qual-supp-sevir} and~\ref{fig:qual-supp-meteo} extend the
qualitative comparison of the main paper. The SEVIR cases are drawn from the same
systematic per-sample CSI-160 ranking as the failure case in the main paper, two
successes and one failure, so the gallery is not hand-picked. The additional
failure is complementary to the one in the main paper. Instead of
under-forecasting a persistent heavy band, the sharp generative head
over-persists a dissipating storm that the baseline correctly fades out.

On MeteoNet, no AlphaPre checkpoint is available (its Table~1 entries are
transcribed from the AlphaPre paper), so the deterministic reference row is our
frozen Lagrangian advection prior, and the panels use the dataset's discrete
reflectivity palette, whose boundaries include the evaluation thresholds
$\{12,18,24,32\}$\,dBZ. The division of labor matches SEVIR. The prior keeps
the system on its observed track but, being a smooth semi-Lagrangian rollout,
dilutes the embedded $32$--$40$\,dBZ cells into the surrounding band as lead time
grows; the generative head restores those heavy cores where the ground truth
keeps them (per-case CSI-32 $0.338$ vs.\ $0.230$ for the prior on index $1138$,
and $0.309$ vs.\ $0.296$ on index $962$). In the failure, the observed scattered
convection keeps regenerating heavy cells across the domain; the prior never
carries them, and the anchored head restores only the strongest core,
under-forecasting the rest. Initiation and re-intensification that the advection
prior cannot represent remain the hardest regime, a limitation inherited from
advection-based nowcasting.

\section*{G\quad Implementation Details}

\input{tex/tables/tab_impl}

Table~\ref{tab:impl} lists the full set of hyperparameters summarized in
Sec.~5.2 of the main paper, grouped by module (optimization, Lagrangian prior,
flow-map head, inference, hardware); symbols are defined in Sec.~4.
The soft-CSI term of the prior objective (Eq.~6 of the main paper) relaxes, for
each evaluation threshold $\kappa$ of the dataset, the hard exceedance mask to a
sigmoid $\sigma((\hat R-\kappa)/\alpha)$ with sharpness $\alpha$, computed on the
dataset's native intensity scale; soft hit, miss, and false-alarm counts are
accumulated, and $1-\mathrm{CSI}_\kappa$ is averaged over the dataset's
thresholds. The warp of Eq.~5 samples on a normalized grid with border padding.

\section*{H\quad Composition-Consistency Ablation}

\input{tex/tables/tab_cc}

Table~\ref{tab:cc} isolates the effect of the composition-consistency term
$\mathcal{L}_{\text{CC}}$ (Eq.~8 of the main paper) on the full model, at its
full training schedule. Removing $\mathcal{L}_{\text{CC}}$ slightly raises the
detection metrics (CSI-M, CSI-181, CSI-219, HSS) but degrades every fidelity and
calibration metric (SSIM, MSE, CRPS). We keep the term enabled for the sharper,
better-calibrated forecasts it produces and do not claim it as a source of
detection skill; the detection gap is small relative to the margins over prior
work reported in the main paper.

\section*{I\quad Structural Verification of the Heavy-Rain Gain}

\input{tex/tables/tab_structural}

\revised{CSI is computed after thresholding and is therefore insensitive to boundary
detail, so a broad forecast could in principle raise it by covering more area.
Table~\ref{tab:structural} tests that alternative explanation directly with the
frequency bias, the ratio of the forecast above-threshold area to the observed
one, together with POD, FAR, the relative area error, and multi-scale FSS.
Area inflation would require a bias above one. Ours is $0.35$ and $0.21$ at
VIL${=}181$ and $219$, so the forecast still covers less heavy-rain area than
the observations, and AlphaPre covers even less ($0.23$ and $0.10$). The gain
therefore appears to be the recovery of heavy rain that blurred forecasts
erase rather than over-prediction, and it is accompanied by the highest POD and the highest FSS
at every neighbourhood scale we evaluated ($1$--$33$\,px). The higher FAR of
the single sample is the expected cost of sharp stochastic detail, and CSI
already accounts for it.

The table also separates the three operating points discussed in
Sec.~4.4 of the main paper. A single sample uses neither the ensemble nor the
probability-matched mean and already exceeds AlphaPre. The direct mean of $16$
samples falls back to the level of AlphaPre at the heavy thresholds, because
averaging cancels displaced peaks, which is the failure mode the
probability-matched mean avoids.}

\section*{J\quad Fusion Ablation: Conditioning versus Additive Residual}

\input{tex/tables/tab_fusion}

\revised{Table~\ref{tab:fusion} completes the controlled grid behind the decoupling
design by crossing the prior content with the mechanism by which the prior
enters the model. With our Lagrangian physics prior the two fusion mechanisms
perform comparably, so we do not claim a skill advantage for conditioning in
isolation. With the blurred deterministic prior, however, only additive
residual fusion degrades, in both heavy-rain skill and calibration. This is
consistent with the residual-anchor failure mode discussed in Sec.~2 of the
main paper, reproduced here under a matched budget: summing a conditional-mean
forecast onto the output re-imposes its blur, whereas conditioning leaves the
head free to replace it. These results indicate that conditioning is the
mechanism robust to the quality of the prior, while the prior content remains
the factor that decides skill, consistent with the prior ablation in the main
paper.}

\section*{K\quad Matched-Ensemble Comparison with a Generative Baseline}

\input{tex/tables/tab_matched}

\revised{The main comparison reports \method{} at its $16$-member operating point
against baselines that are single deterministic forecasts, so we additionally
compare against a generative baseline under a matched ensemble and matched
post-processing. FlowCast (Sec.~2 of the main paper) is absent from Tab.~1 because its released
code and checkpoint cover SEVIR only, at a resolution and horizon that differ
from our protocol, so we reproduced it and adapted it to this protocol.
Generating a full $16$-member ensemble for it takes roughly $72$ hours, so
both methods are evaluated with eight members on a fixed, unselected subset of
the first $512$ test sequences.
In Table~\ref{tab:matched}, \method{} remains ahead in both detection and
calibration while using $2.5$ times fewer network evaluations per member.}

%% file: tex/tables/tab_inference.tex
\begin{table}[!htb]
  \centering
  \caption{Effect of ensemble size on SEVIR (four sampling steps,
  probability-matched mean). Gains per doubling diminish sharply; $16$ members,
  our default, is the accuracy--compute trade-off. In each column, the best is
  \textbf{bold}, the second best \underline{underlined}.}
  \label{tab:inference}
  \setlength{\tabcolsep}{6pt}
  \begin{tabular}{l cccc}
    \toprule
    Ensemble size & CSI-M$\uparrow$ & CSI-181$\uparrow$ & HSS$\uparrow$ & MSE$\downarrow$ \\
    \midrule
    $1$  & 0.3394 & 0.1667 & 0.4335 & 366.80 \\
    $4$  & 0.3559 & 0.1816 & 0.4529 & 327.32 \\
    $8$  & 0.3595 & 0.1843 & 0.4570 & 319.65 \\
    $16$ (default) & 0.3614 & 0.1859 & 0.4593 & 315.67 \\
    $32$ & \underline{0.3623} & \underline{0.1865} & \underline{0.4603} & \underline{313.64} \\
    $64$ & \textbf{0.3627} & \textbf{0.1867} & \textbf{0.4607} & \textbf{312.66} \\
    \bottomrule
  \end{tabular}
\end{table}

%% file: tex/tables/tab_nfe.tex
\begin{table}[!htb]
  \centering
  \caption{Effect of the number of sampling steps $N$ on SEVIR ($16$-member
  probability-matched mean fixed). Detection skill saturates: $N{=}4$ retains
  $99.8\%$ of the CSI-M of $N{=}8$ at half the cost per forecast, and even
  $N{=}1$ exceeds the strongest deterministic baseline (CSI-M $0.3259$).
  In each column, the best is \textbf{bold}, the second best \underline{underlined}.}
  \label{tab:nfe}
  \setlength{\tabcolsep}{6pt}
  \begin{tabular}{l cccc}
    \toprule
    Sampling steps & CSI-M$\uparrow$ & CSI-181$\uparrow$ & HSS$\uparrow$ & MSE$\downarrow$ \\
    \midrule
    $1$ & 0.3463 & 0.1590 & 0.4361 & \textbf{304.50} \\
    $2$ & 0.3523 & 0.1698 & 0.4453 & \underline{307.27} \\
    $4$ (default) & \underline{0.3614} & \underline{0.1859} & \underline{0.4593} & 315.67 \\
    $8$ & \textbf{0.3620} & \textbf{0.1898} & \textbf{0.4616} & 324.98 \\
    \bottomrule
  \end{tabular}
\end{table}

%% file: tex/tables/tab_impl.tex
\begin{table}[!htb]
  \centering
  \caption{Implementation details and hyperparameters, grouped by module.
  Values are given as SEVIR~/~MeteoNet where they differ by dataset; symbols
  are defined in Sec.~4 of the main paper.}
  \label{tab:impl}
  \setlength{\tabcolsep}{6pt}
  \setlength{\aboverulesep}{0pt}\setlength{\belowrulesep}{0pt}
  \renewcommand{\arraystretch}{1.12}
  \begin{tabular}{ll}
    \toprule
    Hyperparameter & Value \\
    \midrule
    \rowcolor{gray!12}
    \multicolumn{2}{l}{\emph{Optimization (both modules)}} \\
    Optimizer & AdamW \\
    Learning rate & $10^{-4}$ \\
    Precision & bf16 \\
    Gradient clipping & $0.5$ (head)~/~$1.0$ (prior) \\
    EMA decay $\rho$ & $0.999$ \\
    \midrule
    \rowcolor{gray!12}
    \multicolumn{2}{l}{\emph{Lagrangian prior}} \\
    Max.\ displacement $d_{\max}$ & $8$ px \\
    Source scale $c_s$ & $0.25$ \\
    Loss weights $(\lambda_{\text{mse}},\lambda_{\text{csi}},\lambda_{v},\lambda_{s})$ & $(0.25,\ 0.02,\ 0.01,\ 10^{-3})$ \\
    Soft-CSI sharpness $\alpha$ & $8$ (native intensity units) \\
    Batch size / warm-up & $16$ / $2$k steps \\
    Training steps & $50$k~/~$30$k \\
    \midrule
    \rowcolor{gray!12}
    \multicolumn{2}{l}{\emph{Flow-map head}} \\
    Base width / channel multipliers & $128$ / $[1,2,3,4]$ \\
    Residual blocks per level & $2$ \\
    Coarsest resolution & $16{\times}16$ (self-attention) \\
    Parameters & $\sim$85M ($\sim$101M with frozen prior) \\
    Min.\ time gap $\delta_{\min}$ & $0.05$ \\
    Direct-map probability $p_0$ & $0.5$ \\
    Composition-consistency weight $\lambda_{\text{CC}}$ & $0\to0.04$, steps $12$k--$30$k \\
    Batch size & $8$ \\
    Warm-up & $5$k steps \\
    Training steps & up to $297.6$k~/~$154$k \\
    \midrule
    \rowcolor{gray!12}
    \multicolumn{2}{l}{\emph{Inference}} \\
    Sampling steps $N$ & $4$ \\
    Ensemble size $K$ & $16$ \\
    \midrule
    \rowcolor{gray!12}
    \multicolumn{2}{l}{\emph{Hardware}} \\
    GPU & 1$\times$ NVIDIA GB202 (96\,GB) \\
    Instance & AWS \texttt{g7e.4xlarge} \\
    Training time (MeteoNet, $154$k steps) & $\sim$11.5\,h \\
    Test evaluation (MeteoNet, $16$-member) & $\sim$7\,min / $1{,}310$ sequences \\
    \bottomrule
  \end{tabular}
\end{table}

%% file: tex/tables/tab_cc.tex
\begin{table}[!htb]
  \centering
  \caption{Composition-consistency ablation on SEVIR (full model, full training
  schedule; $16$-member PMM, four sampling steps). Removing
  $\mathcal{L}_{\text{CC}}$ slightly raises the detection metrics while degrading
  every fidelity metric. In each column, the best result is in \textbf{bold}.}
  \label{tab:cc}
  \setlength{\tabcolsep}{4pt}
  \setlength{\aboverulesep}{0pt}\setlength{\belowrulesep}{0pt}
  \renewcommand{\arraystretch}{1.12}
  \resizebox{\linewidth}{!}{%
  \begin{tabular}{l ccc cc cc}
    \toprule
    Variant & CSI-M$\uparrow$ & CSI-181$\uparrow$ & CSI-219$\uparrow$ & HSS$\uparrow$ & SSIM$\uparrow$ & MSE$\downarrow$ & CRPS$\downarrow$ \\
    \midrule
    without $\mathcal{L}_{\text{CC}}$ & \textbf{0.3652} & \textbf{0.1922} & \textbf{0.0973} & \textbf{0.4646} & 0.7286 & 318.71 & 0.0296 \\
    \rowcolor{blue!8}
    with $\mathcal{L}_{\text{CC}}$    & 0.3614 & 0.1859 & 0.0925 & 0.4593 & \textbf{0.7291} & \textbf{315.67} & \textbf{0.0294} \\
    \bottomrule
  \end{tabular}%
  }
\end{table}

%% file: tex/tables/tab_structural.tex
\begin{table}[!htb]
  \centering
  \caption{Structural verification on the full SEVIR test set
  ($8{,}096$ sequences) at the heavy-rain thresholds VIL${=}181$\,/\,$219$.
  FSS uses a $17$\,px neighbourhood, and the area error is the relative error
  of the above-threshold area. Scores accumulate contingency counts over the
  test set rather than averaging per-sample CSI as in Table~1 of the main
  paper, so the values are comparable across rows here but not with that table.
  In each column, the best result is in \textbf{bold}.}
  \label{tab:structural}
  \setlength{\tabcolsep}{4pt}
  \setlength{\aboverulesep}{0pt}\setlength{\belowrulesep}{0pt}
  \renewcommand{\arraystretch}{1.12}
  \resizebox{\linewidth}{!}{%
  \begin{tabular}{l cccccc}
    \toprule
    Method & CSI$\uparrow$ & Bias$\to 1$ & POD$\uparrow$ & FAR$\downarrow$ & FSS$\uparrow$ & Area err.$\downarrow$ \\
    \midrule
    AlphaPre                    & .157\,/\,.071 & .234\,/\,.104 & .168\,/\,.073 & \textbf{.284}\,/\,.295 & .517\,/\,.258 & .768\,/\,.899 \\
    \method{} (single sample)   & .170\,/\,.084 & .345\,/\,.201 & .196\,/\,.093 & .434\,/\,.534 & .622\,/\,.378 & .660\,/\,.812 \\
    \method{} (mean of $16$)    & .161\,/\,.071 & .240\,/\,.101 & .172\,/\,.073 & .285\,/\,\textbf{.283} & .537\,/\,.258 & .761\,/\,.900 \\
    \rowcolor{blue!8}
    \method{} (PMM, default)    & \textbf{.189}\,/\,\textbf{.097} & \textbf{.349}\,/\,\textbf{.213} & \textbf{.214}\,/\,\textbf{.107} & .387\,/\,.498 & \textbf{.645}\,/\,\textbf{.411} & \textbf{.654}\,/\,\textbf{.797} \\
    \bottomrule
  \end{tabular}%
  }
\end{table}

%% file: tex/tables/tab_fusion.tex
\begin{table}[!htb]
  \centering
  \caption{Fusion ablation on SEVIR. All four variants share the identical
  flow-map head and conditioning channels at a matched $150$k-step budget and
  are evaluated at the same operating point ($16$-member PMM, four sampling
  steps); only the prior content (rows) and the way the prior enters the model
  (columns) change. Cells give CSI-$219\uparrow$\,/\,CRPS$\downarrow$.}
  \label{tab:fusion}
  \setlength{\tabcolsep}{6pt}
  \begin{tabular}{l cc}
    \toprule
    Prior $\backslash$ Fusion & Conditioning (ours) & Additive residual \\
    \midrule
    Lagrangian physics (ours) & .0907\,/\,.0298 & .0931\,/\,.0297 \\
    Deterministic (AlphaPre)  & .0892\,/\,.0299 & .0846\,/\,.0306 \\
    \bottomrule
  \end{tabular}
\end{table}

%% file: tex/tables/tab_matched.tex
\begin{table}[!htb]
  \centering
  \caption{Matched-ensemble comparison on SEVIR. Both methods use eight
  members aggregated by the same probability-matched mean and are scored by
  the same evaluator on a fixed subset of the first $512$ test sequences,
  chosen without selection. NFE is the number of network evaluations per
  member.}
  \label{tab:matched}
  \setlength{\tabcolsep}{6pt}
  \setlength{\aboverulesep}{0pt}\setlength{\belowrulesep}{0pt}
  \renewcommand{\arraystretch}{1.12}
  \begin{tabular}{l ccc}
    \toprule
    Method & NFE\,/\,member & CSI-M$\uparrow$ & CRPS$\downarrow$ \\
    \midrule
    FlowCast (reproduced) & $10$ & $.348$ & $.038$ \\
    \rowcolor{blue!8}
    \method{}                             & $\mathbf{4}$ & $\mathbf{.378}$ & $\mathbf{.031}$ \\
    \bottomrule
  \end{tabular}
\end{table}